\documentclass[10pt,twocolumn]{article}
\usepackage{cvpr}
\usepackage{times}
\usepackage{epsfig}
\usepackage{graphicx}
\usepackage{amsmath}
\usepackage{amssymb}
\usepackage{subcaption}
\usepackage{paralist}
\usepackage[export]{adjustbox}
\usepackage[pagebackref=true,breaklinks=true,letterpaper=true,colorlinks,bookmarks=false]{hyperref}

\cvprfinalcopy

\newcommand{\pa}[1]{\noindent \textbf{#1}. \quad}

\ifcvprfinal\fi 

\begin{document}

\title{Your Local GAN: \\ Designing Two Dimensional Local Attention Mechanisms for Generative Models }

\author{Giannis Daras\\
National Technical University of Athens\\
{\tt\small daras.giannhs@gmail.com}
\and 
Augustus Odena\\
Google Brain\\
{\tt\small augustusodena@google.com}
\and 
Han Zhang\\
Google Brain\\
{\tt\small zhanghan@google.com}
\and 
Alexandros G. Dimakis\\
UT Austin\\
{\tt\small dimakis@austin.utexas.edu }
}

\maketitle

\begin{abstract}
We introduce a new \textit{local sparse attention layer} that preserves  two-dimensional geometry and locality. 
We show that by just replacing the dense attention layer of SAGAN with our construction, we obtain very significant FID, Inception score and pure visual improvements. FID score is improved from 18.65 to \textbf{15.94} on ImageNet, keeping all other parameters the same. 
The sparse attention patterns that we propose for our new layer are designed using a novel information theoretic criterion that uses information flow graphs. 

 We also present a novel way to invert Generative Adversarial Networks with attention. Our method uses the attention layer of the discriminator to create an innovative loss function. This allows us to visualize the newly introduced attention heads and show that they indeed capture interesting aspects of two-dimensional geometry of real images. 
\end{abstract}

\begin{figure*}[!htb]
\begin{center}
    \begin{subfigure}{0.2\textwidth}
     \includegraphics[scale=0.8]{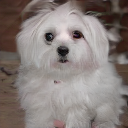}
   \end{subfigure}
    \begin{subfigure}{0.2\textwidth}
     \includegraphics[scale=0.8]{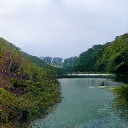}
   \end{subfigure}
    \begin{subfigure}{0.2\textwidth}
     \includegraphics[scale=0.8]{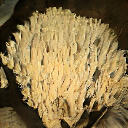}
   \end{subfigure}
    \begin{subfigure}{0.2\textwidth}
     \includegraphics[scale=0.8]{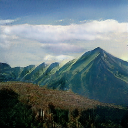}
   \end{subfigure}
\caption{ Samples generated by our model YLG-SAGAN after training on ImageNet. The images are visually significantly better compared to the SAGAN baseline, as also supported by FID and Inception score metrics.}
\end{center}
\end{figure*}

\section{Introduction}

Generative Adversarial Networks~\cite{gans} are making significant progress on modeling and generating natural images~\cite{sagan, biggan}. 
Transposed convolutional layers are a fundmamental architectural component since they capture spatial invariance, a key property of natural images~\cite{dcgan, stylegan, stackgan}.
The central limitation (e.g. as argued in ~\cite{sagan}) is that 
convolutions fail to model complex geometries and long-distance dependencies-- the canonical example is generating dogs with fewer or more than four legs. 


To compensate for this limitation, attention layers~\cite{attention_is_all_you_need} have been introduced in deep generative models~\cite{sagan, biggan}.
Attention enables the modeling of long range spatial dependencies in a single layer
which automatically finds correlated parts of the image even if they are far apart. 
First introduced in SAGAN~\cite{sagan} and further improved in BigGAN~\cite{biggan}, attention layers have led to some of the best known GANs currently available. 

Attention layers have a few limitations. The first is that they are 
computationally inefficient: Standard dense attention requires 
memory and time complexity that scales quadratically in the size of the input. Second, dense attention layers are statistically inefficient: A significant number of training samples is required to train attention layers, a problem that becomes more pronounced when multiple attention heads or layers are introduced~\cite{child2019generating}. Statistical inefficiency also stems from the fact that dense attention does not benefit from locality, since most dependencies in images relate to nearby neighborhoods of pixels. Recent work indicates that most attention layer heads learn to attend mainly to local neighborhoods~\cite{adaptive_span}. 

To mitigate these limitations, sparse attention layers were recently introduced in Sparse Transformers~\cite{child2019generating}. In that paper, different types of sparse attention kernels were introduced and used to obtain excellent results for images, text and audio data. 
They key observation we make is that the patterns that were introduced in Sparse Transformers are actually designed for one-dimensional data, such as text-sequences. 
Sparse Transformers~\cite{child2019generating} were applied to images by reshaping tensors
in a way that significantly distorts distances of the two-dimensional grid of image pixels. 
Therefore, local sparse attention kernels introduced in Sparse Transformers fail to capture image locality. 

\noindent \textbf{Our Contributions:} 
\begin{itemize}
\item We introduce a new 
\textit{local sparse attention layer} that preserves two-dimensional image locality and can support good information flow through attention steps.  

    \item To design our attention patterns we use the information theoretic framework of Information Flow Graphs~\cite{dimakis2010network}. 
    This quantifies how information can flow through multiple steps and preserve two-dimensional locality. 
    We visualize learned attention maps and show that different heads indeed learn different aspects of the geometry of generated images. 

\item We modify SAGAN~\cite{sagan} using our new two-dimensional sparse attention layers to introduce YLG-SAGAN. We empirically show that this change yields significant benefits. We train on ImageNet-128 and we achieve \textbf{14.53\%} improvement to the FID score of SAGAN and \textbf{8.95\%} improvement in Inception score, by only changing the attention layer 
while maintaining all other parameters of the architecture.
Our ablation study shows that indeed the benefits come from two dimensional inductive bias and not from introducing multiple attention heads. 
Furthermore, YLG-SAGAN achieves this performance in $800k$ training steps as opposed to $1300k$ for SAGAN and hence reduces the training time by approximately $40\%$.

\item 
To visualize our attention maps on natural images, we came across the problem of inverting a generator: given an image $x$, how to find a latent code $z$ so that $G(z)$ is as close as possible to $x$. The natural inversion process of performing gradient descent on this loss works in small GANs~\cite{bora2017compressed,rick2017one,raj2019gan,kabkab2018task} but has been notoriously failing in bigger models with attention like SAGAN\footnote{This fact is folklore, known at least among researchers who try to solve inverse problems using deep generative models.
There are, of course numerous other ways to invert, like training an encoder, but also show poor performance on modern GANs with attention. }.  
We present a solution to the GAN inversion problem: We use the attention layer of the discriminator to obtain a weighting on the loss function that subsequently we use to invert with gradient descent. 
%
%
We empirically show excellent inversion results for numerous cases where standard gradient descent inversion fails. 
\end{itemize}


We open-source our code and our models to encourage further research in this area. The code is available under the repository: \href{https://github.com/giannisdaras/ylg}{https://github.com/giannisdaras/ylg} \footnote{The code for our experiments is based on the \href{https://github.com/tensorflow/gan}{tensorflow-gan} library.}

\section{Background}
\pa{Dense Attention}
Given matrices $X \in \mathbb{R}^{N_X \times E_X}, Y \in \mathbb{R}^{N_Y \times E_Y}$, attention of $X$ to $Y$, updates the vector representation of $X$ by integrating the vector representation of $Y$. In this paper, $X, Y$ are intermediate image representations. More specifically, attention of $X$ to $Y$ associates the following matrices with the inputs: The key matrix $X_K = X \cdot W_K$, the query matrix $Y_Q = Y \cdot W_Q$ and the value matrix $Y_V = X \cdot W_V$ where $W_K \in \mathbb{R}^{E_X \times E}, W_Q \in \mathbb{R}^{E_Y \times E}, W_V \in \mathbb{R}^{E \times E_V}$ are learnable weight matrices. Intuitively, queries are compared with keys 
and values translate the result of this comparison to a new vector representation of $X$ that integrates information from $Y$.
Mathematically, the output of the attention is the matrix:
$X' = \sigma( X_Q \cdot Y_K^T) \cdot Y_V$ where $\sigma(\cdot)$ denotes the softmax operation along the last axis.

\pa{Sparsified Attention}
\label{bg_attention}
The quadratic complexity of attention to the size of the input is due to the calculation of the matrix $A_{X, Y} = X_Q \cdot Y_K^T, \in \mathbb{R}^{N_X \times N_Y}$. Instead of performing this calculation jointly, we can split attention in multiple steps. At each step $i$, we attend to a subset of input positions, specified by a binary mask $M_i \in \{0, 1\}^{N_X \times N_Y}$. Mathematically, at step $i$ we calculate matrix $A^{i}_{X, Y}$, where: $A^{i}_{X, Y}[a, b] = \begin{cases}
A_{X, Y}[a, b], \quad M^i[a, b] = 1 \\
-\infty, \quad M^i[a, b] = 0
\end{cases}$. 
\linebreak In this expression, $-\infty$ means that after the softmax, this position will be zeroed and thus not contribute to the calculation of the output matrix.
The design of the masks $\{M^i\}$ is key in reducing the number of positions attended. 

There are several ways that we can use the matrices $A^{i}_{X, Y}$ to perform multi-step attention~\cite{child2019generating} in practice.
The simplest is to have separate attention heads~\cite{attention_is_all_you_need} calculating the different matrices
$\{A^i_{X, Y}\}$ in parallel and then concatenate along the feature dimension. We use this approach in this paper.

\section{Your Local GAN}
\subsection{Full Information Attention Sparsification}
\label{full_information}
As explained, an attention sparsification in $p$ steps is described by binary masks $\{M^1, ..., M^p\}$.
The question is how to design a good set of masks for these attention steps. 
We introduce a tool from information theory to guide this design. 

Information Flow Graphs are directed acyclic graphs introduced
in~\cite{dimakis2010network} to model distributed storage systems through network information flow~\cite{ahlswede2000network}.
For our problem, this graph models how information flows across attention steps. For a given set of masks $\{M^1, ..., M^p\}$, we create a multi-partite graph $G(V= \{V^0, V^1, ..., V^p\}, E)$ where directed connections
between $V^i, V^{i+1}$ are determined by mask $M^i$.
Each group of vertices in partition $V^i$ corresponds to attention tokens of step $i$. 

\begin{figure*}[!htb]
    \begin{center}
            \begin{subfigure}{0.31\textwidth}
            \centering
            \includegraphics[scale=0.6]{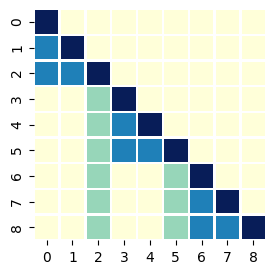}
            \subcaption{\footnotesize{Attention masks for Fixed Pattern~\cite{child2019generating}.} }
            \label{ST_mask}
        \end{subfigure}
        \quad
        \begin{subfigure}{0.31\textwidth}
            \centering
            \includegraphics[scale=0.6]{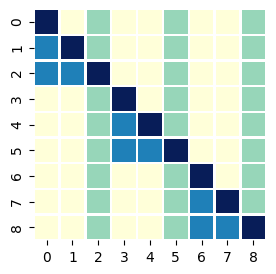}
            \subcaption{\footnotesize{Attention masks for Left To Right (LTR) pattern.}}
            \label{LTR_mask}
        \end{subfigure}
        \quad
        \begin{subfigure}{0.31\textwidth}
            \centering
            \includegraphics[scale=0.6]{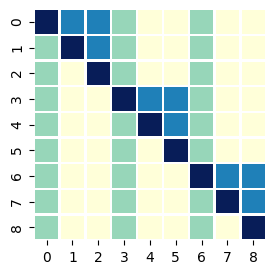}
            \subcaption{\footnotesize{Attention masks for Right To Left (RTL) pattern.}}
            \label{RTL_mask}
        \end{subfigure}
        
        \begin{subfigure}{0.31\textwidth}
            \centering
            \includegraphics[scale=0.60]{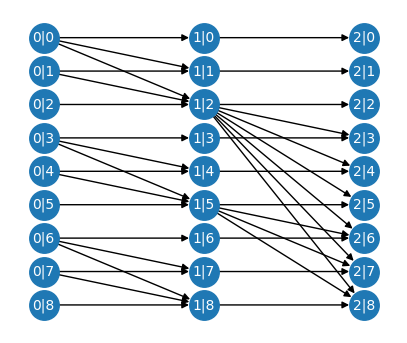}
            \subcaption{\footnotesize{Information Flow Graph associated with Fixed Pattern. This pattern \textit{does not have Full Information}, i.e. there are dependencies between nodes that the attention layer cannot model. For example, there is no path from node $0$ of $V^0$ to node $1$ of $V^2$.}}
            \label{ST_IFG}
        \end{subfigure}
        \quad
        \begin{subfigure}{0.31\textwidth}
            \centering
            \includegraphics[scale=0.60]{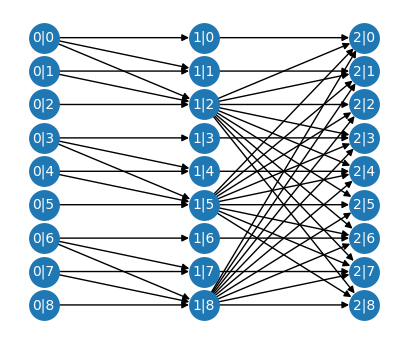}
            \subcaption{
            \footnotesize{
            Information Flow Graph associated with LTR. This pattern has \textbf{Full Information}, i.e. there is a path between any node of $V^0$ and any node of $V^2$. Note that the number of edges is only increased by a constant compared to the Fixed Attention Pattern~\cite{child2019generating}, illustrated in \ref{ST_IFG}.}}
            \label{LTR_IFG}
        \end{subfigure}
        \quad
        \begin{subfigure}{0.31\textwidth}
            \centering
            \includegraphics[scale=0.6]{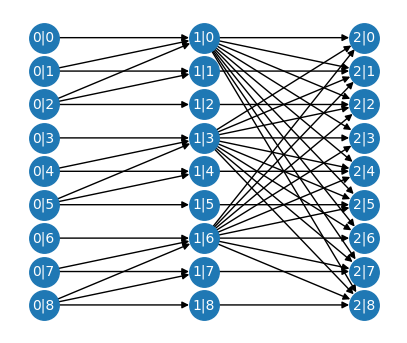}
            \subcaption{\footnotesize{Information Flow Graph associated with RTL. This pattern also has \textbf{Full Information}. RTL is a "transposed" version of LTR, so that local context at the right of each node is attended at the first step.}}
            \label{RTL_IFG}
        \end{subfigure}
    \end{center}
    \caption{\small{This Figure illustrates the different 2-step sparsifications of the attention layer we examine in this paper. First row demonstrates the different boolean masks that we apply to each of the two steps. Color of cell [i. j] indicates whether node i can attend to node j. With dark blue we indicate the attended positions in both steps. With light blue the positions of the first mask and with green the positions of the second mask. The yellow cells correspond to positions that we do not attend to any step (sparsity). The second row illustrates Information Flow Graph associated with the aforementioned attention masks. An Information Flow Graph visualizes how information "flows" in the attention layer. Intuitively, it visualizes how our model can use the 2-step factorization to find dependencies between image pixels. At each multipartite graph, the nodes of the first vertex set correspond to the image pixels, just before the attention. An edge from a node of the first vertex set, $V^0$, to a node of the second vertex set, $V^1$, means that the node of $V^0$ can attend to node of $V^1$ at the first attention step. Edges between $V^1, V^2$ illustrate the second attention step.}}
    \label{patterns}
\end{figure*}

We say that an attention sparsification has \textbf{Full Information} if its corresponding Information Flow Graph has a directed path from every node
$a \in V^0$ to every node $b \in V^p$.
Please note that the Fixed pattern~\cite{child2019generating} shown in sub-figure \ref{ST_mask} does not have Full Information:
there is no path from node 1 of $V^0$ to node 2 of $V^2$.

Sparse attention is usually considered as a way to reduce the computational overhead of dense attention at a hopefully small performance loss. 
However, we show that attention masks chosen with a bias toward two-dimensional locality, can surprisingly \textit{outperform} dense attention layers (compare the second and the third row of Table \ref{results}).
This is an example of what we call the statistical inefficiency 
of dense attention. Sparse attention layers with locality create better inductive bias and hence can perform better in the finite sample regime. In the limit of infinite data, dense attention can always simulate sparse attention or perform better, in the same way that a fully connected layer can simulate a convolutional layer for a possible selection of weights.

We design the sparse patterns of YLG as the natural extensions of the patterns of~\cite{child2019generating} while ensuring that the corresponding Information Flow Graph supports Full Information. 
The first pattern, which we call Left to Right (LTR), extends the pattern of~\cite{child2019generating} to a bi-directional context.
The second pattern, which we call Right to Left (RTL), is a transposed version of LTR.
The corresponding $9 \times 9$ masks and associated Information Flow Graphs are presented in sub-figures \ref{LTR_mask}, \ref{LTR_IFG} (LTR) and \ref{RTL_mask}, \ref{RTL_IFG} (RTL).
These patterns allow attention only to $n\sqrt n$ positions, significantly reducing the quadratic complexity of dense attention.
It is possible to create very sparse Full Information graphs using multiple attention steps, but designing them and training them remains open for future work; in this paper we focus on two-step factorizations. We include more details about information flow graphs and how we use them to design attention patterns in the Appendix.

\subsection{Two-Dimensional Locality}
\label{grid_aware}
\begin{figure}[!htb]
    \begin{center}
        \begin{subfigure}{0.22\textwidth}
        \includegraphics[scale=0.6]{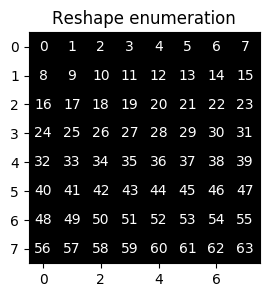}
        \label{fig:reshape_enum}
        \end{subfigure}
        \quad
        \begin{subfigure}{0.22\textwidth}
        \centering
            \includegraphics[scale=0.6]{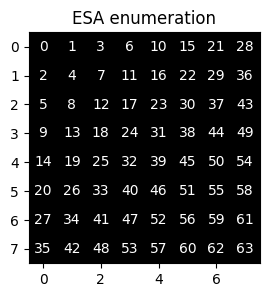}
            \label{fig:esa_enum}
        \end{subfigure}
    \end{center}
    \vspace{-0.5cm}
    \caption{\footnotesize{
    Reshape and ESA enumerations of the cells of an image grid that show 
    how image grid is projected into a line.
    (Left) Enumeration of pixels of an $8\times8$ image using a standard reshape.
    This projection maintains locality only in rows.
    (Right)  Enumeration of pixels of an $8\times8$ image, using the ESA framework. 
    We use the Manhattan distance from the start $(0, 0)$ as a criterion for enumeration.
    Although there is some distortion due to the projection into 1-D, locality is mostly maintained.
    }}
    \label{fig:enumerations}
\end{figure}

The factorization patterns of Sparse Transformers~\cite{child2019generating} and their Full Information extensions illustrated in 
Figure \ref{patterns} are 
fundamentally matched to one-dimensional data, such as text-sequences. 

The standard way to apply these layers on images is to reshape the three dimensional image tensors (having three color channels) to a two-dimensional tensor 
$X \in R^{N \times C}$ that enters attention. This corresponds to $N$ tokens, each containing a $C$-dimensional representation of a region of the input image. 
This reshape arranges these $N$ tokens linearly, significantly distorting which parts of the image are nearby in two dimensions. 
This behavior is illustrated in the sub-figure at the left of Figure \ref{fig:enumerations}. 

We argue that this is the reason that one-dimensional sparsifications are not ideal for images. In fact, the authors of \cite{child2019generating} mention that the Fixed Pattern (Figure \ref{ST_mask})
was designed for text-sequences and not for images. 
Our central finding is that these patterns \textit{can} work very well for images, if their two dimensional structure is correctly considered. 

The question is therefore how to take two-dimensional locality into account. 
We could create two-dimensional attention patterns directly on a grid 
but this would have significant computational overhead and also prevent us from extending one dimensional sparsifications that are known to work well~\cite{star,child2019generating}. 
Instead, we modify one dimensional sparsifications to become aware of two-dimensional locality with the following trick:  (i) we enumerate pixels of the image based on their Manhattan distance from the pixel at location (0, 0) (breaking ties using row priority), (ii) shift the indices of any given one-dimensional sparsification to match the Manhattan distance enumeration instead of the reshape enumeration, and (iii) apply this new one dimensional sparsification pattern, that respects two-dimensional locality, to the one-dimensional reshaped version of the image.
We call this procedure ESA (Enumerate, Shift, Apply) and illustrate it in Figure \ref{fig:enumerations}.

The ESA trick introduces some distortion compared to a true two-dimensional distance. We found however that this was not too limiting, at least for $128 \times 128$ resolution. 
On the other hand, ESA offers an important implementation advantage: it theoretically allows the use of one-dimensional block-sparse kernels~\cite{BLOCKSPARSE}. 
Currently these kernels exist only for GPUs, but making them work for TPUs is
still under development. 

\section{Experimental Validation}
\label{experiments}
We conduct experiments on the challenging ImageNet~\cite{ILSVRC15} dataset. 
We choose SAGAN~\cite{sagan} as the baseline for our models because, unlike BigGAN~\cite{biggan} it has official open-source Tensorflow code.
BigGAN is not open-source and therefore training or modifying this architecture was not possible\footnote{Note that there is an `unofficial' BigGAN that is open in PyTorch. However, that implementation uses gradient checkpointing and requires $8$ V100 GPUS for $15$ days to train. We simply did not have such computing resources. We believe, however, that YLG can be easily combined with BigGAN (by simply replacing its dense attention layer) and will yield an even better model.}.

In all our experiments, we change \textit{only the attention layer} of SAGAN, keeping all the other hyper-parameters unchanged (the number of parameters is not affected). We trained all models for up to 1,500,000 steps on individual Cloud TPU v3 devices (v3-8) using a $1e^{-4}$ learning rate for generator and $4e^{-4}$ for the discriminator. For all the models we report the best performance obtained, even if it was obtained at an earlier point during training. 

\paragraph{Attention Mechanism}
We start with the Fixed Pattern (Figure \ref{ST_mask}) and modify it:
First, we create Full Information extensions (Section \ref{full_information}), 
yielding the patterns Left-To-Right (LTR) and Right-To-Left (RTL) (Figures \ref{LTR_mask} and \ref{RTL_mask} respectively).
We implement multi-step attention in parallel using different heads. 
Since each pattern is a two-step sparsification, this yields 4 attention heads.
To encourage diversity of learned patterns, we use each pattern twice, so the total number of heads in our new attention layer is 8.
We use our ESA procedure (Section \ref{grid_aware}) to render these patterns aware of two dimensional geometry.

\paragraph{Non-Square Attention} 
In SAGAN, the query image and the key image in the attention layer have different dimensions. 
This complicates things, because the sparsification patterns we discuss are designed for self-attention,
where the number of query and key nodes is the same.
Specifically, for SAGAN the query image is $32\times 32$ and the key image is $16\times 16$.
We deal with this in the simplest possible way:
we create masks for the $16\times 16$ image and we shift these masks to cover the area of the $32\times 32$ image. 
Thus every $16 \times 16$ block of the $32\times 32$ query image attends with full information to the $16\times 16$ key image.

\begin{table}[!htb]
\begin{tabular}{l|l|l|l|l|}
      & \# Heads & FID            & Inception      \\ \hline
SAGAN & 1        & 18.65          & 52.52          \\
SAGAN & 8 & 20.09  & 46.01 \\
\textbf{YLG-SAGAN}   & 8        & \textbf{15.94} & \textbf{57.22} \\
YLG - No ESA & 8 & 17.47 & 51.09  \\
YLG - Strided & 8 & 16.64 & 55.21 \\
\end{tabular}
\caption{\footnotesize ImageNet Results: Table of results after training SAGAN and YLG-SAGAN on ImageNet. Table also includes Ablation Studies (SAGAN 8 heads, YLG - No ESA, YLG - Strided). Our best model, \textbf{YLG}, achieves \textbf{15.94} FID and \textbf{57.22} Inception score. Our scores correspond to \textbf{14.53\%} and \textbf{8.95\%} improvement to FID and Inception respectively.
We emphasize that these benefits are obtained by only one layer change to SAGAN, replacing dense attention with the local sparse attention layer that we introduce.}
\label{results}
\end{table}

\noindent  \textbf{Results:}
As shown in Table \ref{results}, YLG-SAGAN (3rd row) outperforms SAGAN by a large margin measured by both FID and Inception score.
 Specifically, YLG-SAGAN increases Inception score to \textbf{57.22} ($8.95\%$ improvement) and improves FID to \textbf{15.94} ($14.53\%$ improvement). Qualitatively, we observe really good-looking samples for categories with simple geometries and homogeneity. Intuitively, a two-dimensional locality can benefit importantly categories such as valleys or mountains, because usually the image transitions for these categories are smoother compared to others and thus the dependencies are mostly local.

\begin{figure*}[!htb]
    \begin{subfigure}{0.44\textwidth}
    \centering
        \includegraphics[scale=0.5]{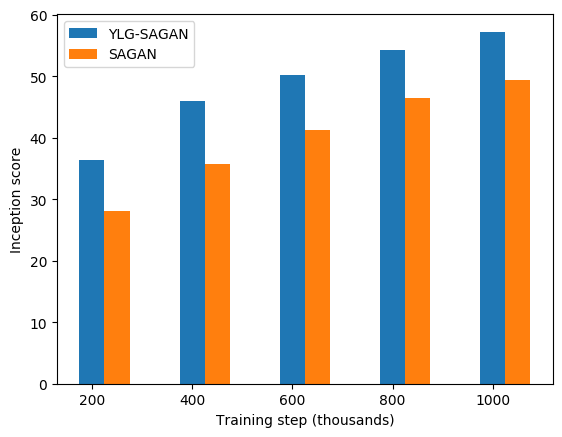}
    \end{subfigure}
    \quad
    \begin{subfigure}{0.44\textwidth}
    \centering
        \includegraphics[scale=0.5]{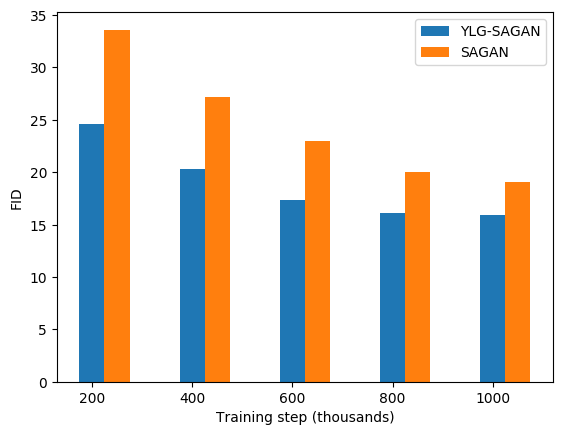}
    \end{subfigure}
    \vspace{-0.2cm}
    \caption{  \footnotesize{ Training comparison for YLG-SAGAN and SAGAN.
    We plot every 200k steps the Inception score (a) and the FID (b) of both YLG-SAGAN and SAGAN, up to 1M training steps on ImageNet. 
    As it can be seen, YLG-SAGAN converges much faster compared to the baseline. Specifically, we obtain our best FID at step 865k, while SAGAN requires over 1.3M steps to reach its FID performance peak.
    Comparing peak performance for both models, we obtain an improvement from $18.65$ to $\mathbf{15.94}$ FID, by only changing the attention layer.
    }}
    \label{logs}
\end{figure*}

Additionally to the significantly improved scores, one important benefit of using YLG sparse layer instead of a dense attention layer, is that we observe significant reduction of the training time needed for the model to reach it's optimal performance. SAGAN reached it's best FID score after more that 1.3 million training steps while YLG-SAGAN reaches its' optimal score after \textbf{only 865,000 steps} ($\approx 40\%$ reduction to the training time). Figure \ref{logs} illustrates SAGAN and YLG-SAGAN FID and Inception score as a function of the training time.  

We create two collages to display samples from our YLG version of SAGAN. At the Upper Panel of Figure \ref{samples}, we show dogs of different breeds generated by our YLG-SAN. At the Lower Panel, we use YLG-SAGAN to generate samples from randomly chosen classes of the ImageNet dataset.

\subsection{Ablation Studies}

\paragraph{Number of Attention Heads}
The Original SAGAN implementation used a single-headed attention mechanism. 
In YLG, we use multiple heads to perform parallel multi-step sparse attention.
Previous work has shown that multiple heads increased performance for Natural Language Processing tasks~\cite{attention_is_all_you_need}.
To understand how multiple heads affect SAGAN performance, we train an 8 head version of SAGAN.
The results are reported in the second row of Table \ref{results}. Multiple heads actually \textbf{worsen} significantly the performance of the original SAGAN, reducing Inception score from 52.52 to 46.01.
We provide a post-hoc interpretation of this result.
The image embedding of the query vector of SAGAN has only 32 vector positions. By using 8 heads, each head gets only 4 positions for its' vector representation. Our intuition is that a 4-positions vector representation is not sufficient for effective encoding of the image information for a dense head and that accounts for the decrease in performance. It is important to note that YLG-SAGAN does not suffer from this problem. The reason is that each head is sparse, which means that only attends to a percentage of the positions that dense head attends to. Thus, a smaller vector representation does not worsen performance. Having multiple divergent sparse heads allows YLG layer to discover complex dependencies in the image space throughout the multi-step attention.

\paragraph{Two-Dimensional Locality}
As described in Section \ref{grid_aware} YLG uses the ESA procedure, to adapt 1-D sparse patterns to data with 2-D structure. Our motivation was that grid-locality could help our sparse attention layer to better model local regions.
In order to validate this experimentally, we trained a version of YLG \textit{without} the ESA procedure.
We call this model YLG - No ESA.
The results are shown in 4th row of Table \ref{results}: without the ESA procedure, the performance of YLG is about the same with the original SAGAN. This experiment indicates that ESA trick is essential for using one-dimensional sparse patterns for grid-structured data. If ESA framework is used, FID improves from $17.47$ to $15.94$ and Inception score from $51.09$ to $57.22$, without any other difference in the model architecture. Thus, ESA is a plug-and-play framework that achieves great performance boosts to both FID and Inception score metrics. ESA allows the utilization of fast sparse one-dimensional patterns that were found to work well for text-sequences to 
be adapted to images, with great performance benefits. In section \ref{inversion_lens}, we visualize attention maps to showcase how our model utilizes ESA framework in practice.

\paragraph{Sparse Patterns}
Our YLG layer uses the LTR and RTL patterns (Figures \ref{LTR_mask} and \ref{RTL_mask} respectively).
Our intuition is that using multiple patterns at the same time increases performance because the model will be able to discover dependencies using
multiple different paths. 
To test this intuition, we ran an experiment using the Full Information extension of the Strided~\cite{child2019generating} pattern. We choose this pattern because it was found to be effective for modeling images~\cite{child2019generating} due to its' periodic structure. As with LTR and RTL patterns, we extend the Strided pattern so that it has Full Information\footnote{We include visualizations of the Full Information Strided Pattern in the Appendix.}. We refer to the YLG model that instead of LTR and RTL patterns, has 8 heads implementing the Strided pattern as YLG - Strided. For our experiment, we use again the ESA trick.
We report the results on the 5th row of Table \ref{results}. YLG - Strided importantly surpasses SAGAN both in FID and Inception score, however, it is still behind YLG. Although in the Sparse Transformers~\cite{child2019generating} it has been claimed that strided pattern is more suitable for images than the patterns we use in YLG, this experiment strongly suggests that it is the grid-locality which makes the difference, as both models are far better than SAGAN. Also, this experiment indicates that multiple sparse patterns can boost performance compared to using a single sparse pattern. To be noted, using multiple different patterns at the same attention layer requires scaling the number of heads as well. Although YLG variations of SAGAN were not impacted negatively by the increase of attention heads, more severe up-scaling of the number of heads could potentially harm performance, similarly to how 8 heads harmed performance of SAGAN.

\section{Inverting Generative Models with Attention}

We are interested in visualizing our sparse attention on real images, not just generated ones.
This leads naturally to the problem of \textit{projecting an image on the range of a generator}, also called inversion. Given a real image $x \in \mathbb{R}^n$ and a generator $G(z)$, inversion corresponds to finding 
a latent variable $z^* \in \mathbb{R}^k$, so that $G(z^*) \in \mathbb{R}^n$ approximates the given image $x$ as well as possible. 
One approach for inversion is to try to solve the following non-convex optimization problem:
\begin{equation}
    \underset{z^*}{\textrm{argmin}}\{ \lVert G(z^*) - x \rVert ^2 \}.
    \label{direct}
\end{equation}

To solve this optimization problem, we can perform gradient descent from a random initalization $z_0$ to minimize this projection distance in the latent space. This approach was introduced independently in several papers~\cite{lipton2017precise,bora2017compressed,rick2017one} and further generalized to solve inverse problems beyond inversion~\cite{bora2017compressed,rick2017one,raj2019gan,kabkab2018task}.
Very recent research~\cite{hand2019global,song2019surfing} demonstrated that for fully connected generators with random weights and sufficient layer expansion, gradient descent will provably converge to the correct optimal inversion. 

Unfortunately, this theory does not apply for generators that have attention layers. Even empirically, inversion by gradient descent fails for bigger generative models like SAGAN and YLG-SAGAN. As we show in our experiments the optimizer gets trapped in local minimima producing reconstructions that only vaguely resemble the target image.
Other approaches for inversion have been tried in the literature, like training jointly an encoder~\cite{donahue2016adversarial} but none of these methods have been known to successfully invert complex generative models with attention layers. 

We propose a novel inversion method that uses the discriminator to solve the minimization problem in  an different representation space. Interestingly, the discriminator yields representations with a smoother loss landscape, especially if we use the attention layer in a special way. In more detail:
We begin with a random latent variable $z$ and a given real image $x$.
We denote with $D^0$ the Discriminator network up to, but not including, the attention layer and obtain the representations $D^0(G(z))$ and $D^0(x)$.
We could perform gradient descent to minimize the distance of these discriminator representations:
    \[ 
        \lVert D^0(G(z)) - D^0(x) \rVert ^2.
    \]

We found, however, that we can use the attention map of the real image to further enhance inversion.
We will use the example of the SAGAN architecture to illustrate this.
Inside the SAGAN Discriminator's attention, an attention map 
$M\in \mathbb R ^{32 \times 32 \times 16 \times 16}$ is calculated.
For each pixel of the $32 \times 32$ image, this attention map is a distribution over the pixels 
of the $16 \times 16$ image. 
We can use this attention map to extract a \textit{saliency map}.
For each pixel of the $16\times 16$ image, we can average the probabilities from all the pixels of the
$32\times32$ image and create a probability distribution of shape $16\times 16$.
We denote this distribution with the letter $S$. 
Intuitively, this distribution represents how important each pixel of the image is to the discriminator.

Our proposed inversion algorithm is to perform gradient descent to minimize the discriminator embedding distance, weighted by these saliency maps: 
\begin{equation}
            \lVert \left( D^0(G(z)) - D^0(x) \right) \cdot S' \rVert ^2,
            \label{disc_loss}
\end{equation}
where $S'$ is a projected version of saliency map $S$ to the dimensions of $D^0(x)$.
We actually calculate one saliency map $S'$ per head and use their sum  as the final loss function that we optimize for inversion. More details are included in the Appendix.

\subsection{Inversion as lens to attention}
\label{inversion_lens}
Given an arbitrary real image, we can now solve for a $z$ yielding
a similar generated image from the generator, and visualize the attention maps.
\begin{figure*}[!htb]
\centering
        \begin{subfigure}{0.15\textwidth}
            \includegraphics[scale=0.6]{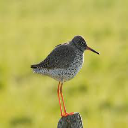}
            \subcaption{}
            \label{redshank}
        \end{subfigure}
        \quad
        \begin{subfigure}{0.15\textwidth}
            \includegraphics[scale=0.6]{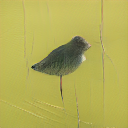}
            \subcaption{}
            \label{redshank_inv_prev}
        \end{subfigure}
        \quad
        \begin{subfigure}{0.15\textwidth}
            \includegraphics[scale=0.6]{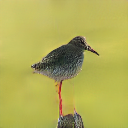}
            \subcaption{}
            \label{redshank_inv}
        \end{subfigure}
        \label{fig:indigo}
        \quad
        \begin{subfigure}{0.15\textwidth}
            \includegraphics[scale=0.29]{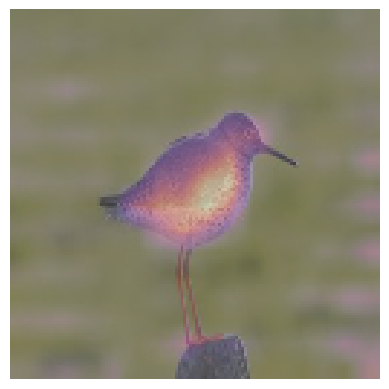}
            \subcaption{}
            \label{sal_7}
        \end{subfigure}
        \quad
        \begin{subfigure}{0.15\textwidth}
            \adjustimage{scale=0.29}{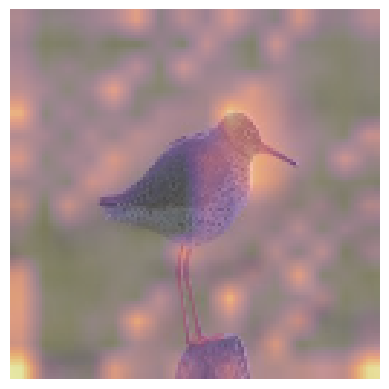}
            \subcaption{}
            \label{sal_2}
        \end{subfigure}
    \vspace*{-3mm}

    \caption{\footnotesize{    Inversion and Saliency maps for different heads of the Generator network.
    We emphasize that this image of a redshank bird was not in the training set, it is rather obtained by a Google image search. 
    Saliency is extracted by averaging the attention each pixel of the key image gets from the query image.
    We use the same trick to enhance inversion.
    (a) A real image of a redshank.
    (b) A demonstration of how the standard inversion method~\cite{bora2017compressed} fails. 
    (c) The inverted image for this redshank, using our technique.
    (d) Saliency map for head 7. Attention is mostly applied to the bird body.
    (e) Saliency map for head 2. This head attends almost everywhere in the image.
    }}
    \label{fig:saliencies}
\end{figure*}

\begin{figure*}[!htb]
        \begin{subfigure}{0.155\textwidth}
            \includegraphics[scale=0.6]{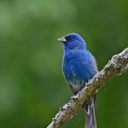}
            \subcaption{}
            \label{indigo}
        \end{subfigure}
        \begin{subfigure}{0.155\textwidth}
            \adjustimage{scale=0.6}{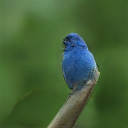}
            \subcaption{}
            \label{indigo_inv_prev}
        \end{subfigure}
        \begin{subfigure}{0.155\textwidth}
            \includegraphics[scale=0.6]{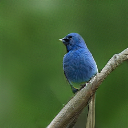}
            \subcaption{}
            \label{indigo_inv}
        \end{subfigure}
        \begin{subfigure}{0.155\textwidth}
            \includegraphics[scale=0.206]{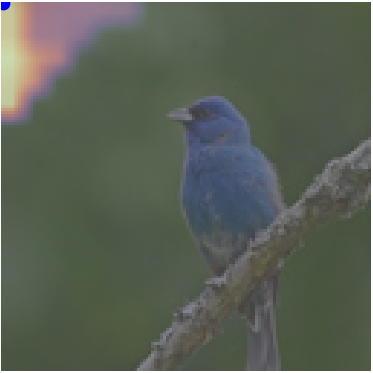}
            \subcaption{}
            \label{manhattan}
        \end{subfigure}
        \begin{subfigure}{0.155\textwidth}
            \adjustimage{scale=0.206}{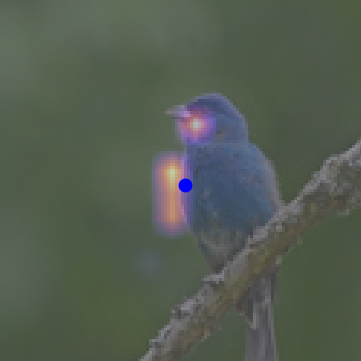}
            \subcaption{}
            \label{local_attn}
        \end{subfigure}
        \begin{subfigure}{0.155\textwidth}
            \adjustimage{scale=0.206}{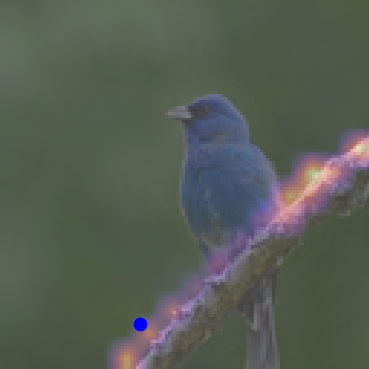}
            \subcaption{}
            \label{non_local}
        \end{subfigure}
        \caption{\footnotesize{
        Inverted image of an indigo bird and visualization of the attention maps for specific query points.
        (a) The original image. Again, this was obtained with a Google image search and was not in the training set. 
        (b) Shows how previous inversion methods fail to reconstruct the head of the bird and the branch.
        (c) A successful inversion using our method.
        (d) Specifically, \ref{manhattan} shows how attention uses our ESA trick to model background, homogeneous areas.
        (e) Attention applied to the bird.
        (f) Attention applied with a query on the branch. Notice how attention is non-local and captures the full branch. 
        }} 
        \label{heads}
\end{figure*}

We explain our approach using an example of a real image of a redshank (Figure \ref{redshank}).  
Figure \ref{redshank_inv_prev} shows how the standard method for inverting
generators~\cite{bora2017compressed} fails: the beak, legs, and rocks are missing.
Figure \ref{redshank_inv} shows the result of our method. 
Using the $z$ that we found using inversion, we can project maps of the attention layer back to the original image to get valuable insight into how the YLG layers work.

First, we analyze the differences between the YLG-SAGAN attention heads.
For each attention head of the generator, we create a saliency map as
described above and use these maps to analyze the attention mechanism.
As shown in Figure \ref{sal_7}, the head-7 in the generator is mostly ignoring background focusing on the bird.
Other heads function differently: The saliency map of head-2 
(Figure \ref{sal_2}) shows that this head attends globally.
We also find that there are heads that that attend quite sparsely, for example,  head-5 attends only to 5-6 background pixels.

We present a second inversion, this time an indigo bird (Figure \ref{indigo}).
Figure \ref{indigo_inv_prev} shows how the standard method
~\cite{bora2017compressed} for inverting fails: the head of the
bird and the branch are not reconstructed. 
We also illustrate where specific query points attend to. 
We first illustrate that the the model exploited the local bias of ESA:
We plot the attention map for query point $(0, 0)$ for generator-head-0.
This point, indicated with a blue dot, is part of the background.
We clearly see a local bias in the positions this point attends to.
Another example of \textit{two-dimensional} local attention is shown in Figure \ref{local_attn}.
This figure illustrates the attention map of generator-head-4 
for a query point on the body of the bird (blue dot).
This point attends to the edges of the bird body and to the bird head.

Finally, Figure \ref{non_local} shows that there are query points that attend to
long-distance, demonstrating that the attention mechanism is capable of exploiting both locality and long-distance relationships when these appear in the image.

\section{Related Work}

\label{related_work}

There has been a flourishing of novel ideas on making attention mechanisms more efficient. 
Dai et al.~\cite{transformer_xl} separate inputs into chunks and associate a state vector with previous chunks of the input. 
Attention is performed per chunk, but information exchange between chunks is possible via the state vector.
Guo et al.~\cite{star} show that a star-shaped topology can reduce attention cost from $O(n^2)$ to $O(n)$ in text
sequences. Interestingly, this topology does have full information, under our framework. 
Sukhbaatar et al.~\cite{adaptive_span} introduced the idea of a learnable adaptive span for each attention layer.
Calian et al.~\cite{scram} proposed a fast randomized algorithm that exploits spatial coherence and sparsity to design sparse approximations.
We believe that all these methods can be possibly combined with YLG, but so far nothing has been demonstrated to improve generative models in a plug-and-play way that this work shows. 

There is also prior work on using attention mechanisms to model images:
One notable example is Zhang et al.~\cite{sagan}, which we have discussed extensively and which adds 
a self-attention mechanism to GANs.
See also Parmar et al.~\cite{imagetransformer}, which uses local-attention that is not multi-step.

\section{Conclusions and Future Work}

We introduced a new type of local sparse attention layer designed for two-dimensional data. 
We believe that our layer will be widely applicable for any model with attention that works on two-dimensional data. 
An interesting future direction is the design of attention layers, thought of as multi-step networks. The two conflicting objectives are to make these networks as sparse as possible (for computational and statistical efficiency) but also support good information flow. We introduced information flow graphs as a mathematical abstraction and proposed full information as a desired criterion for such attention networks. 

Finally, we presented a novel way to solve the inversion problem for GANs. Our technique uses the discriminator in two ways: First, using its attention to obtain pixel importance and second, as a smoothing representation of the inversion loss landscape. 
This new inversion method allowed us to visualize our network on approximations of real images and also to test how good a generative model is in this important coverage task. We believe that this is the first key step towards using generative models for inverse problems and we plan to explore this further in the future.

\begin{figure}[!htb]
\begin{subfigure}{0.45\textwidth}
\centering
\includegraphics[scale=0.07]{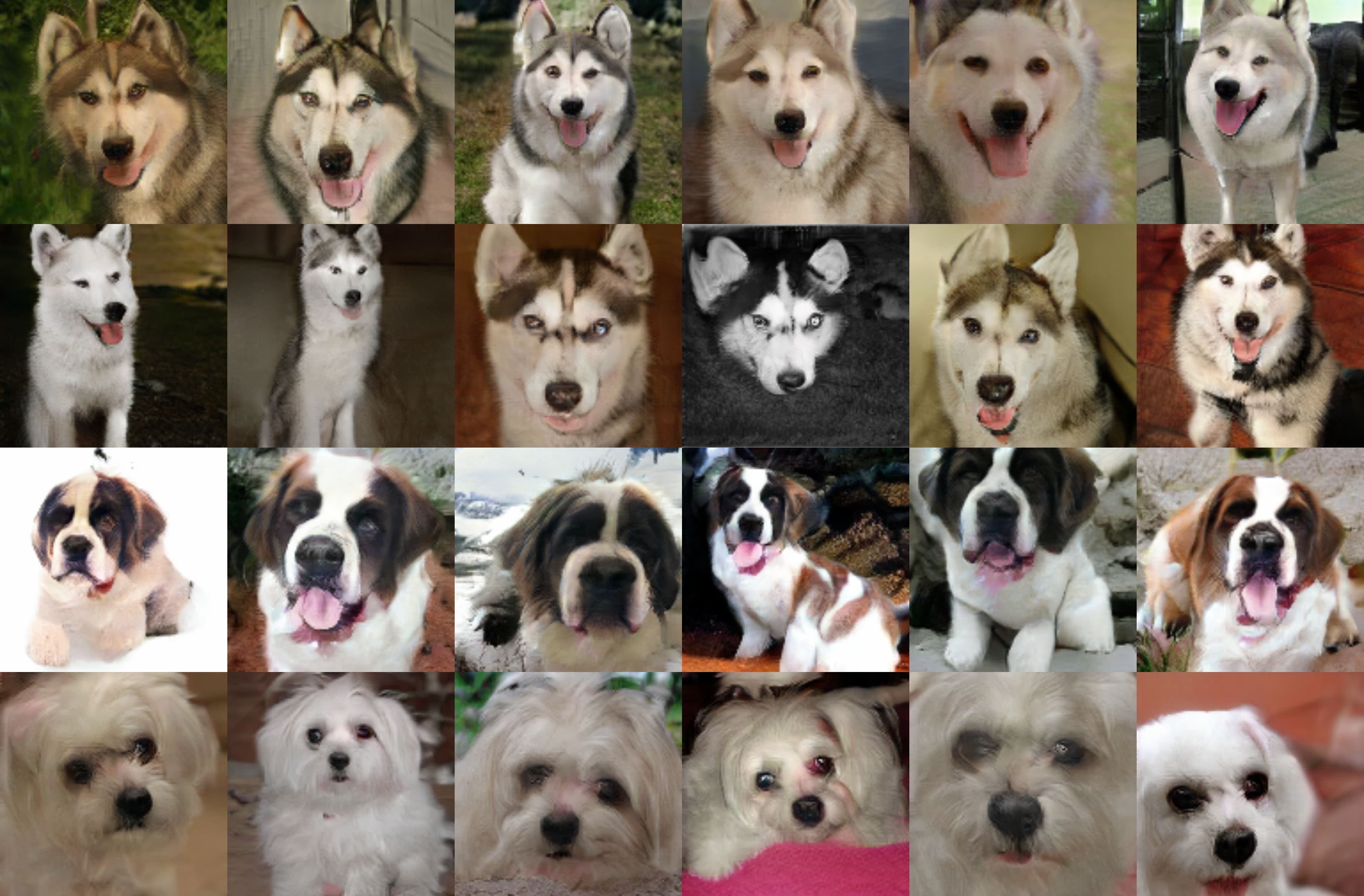}
\end{subfigure}
\begin{subfigure}{0.45\textwidth}
\centering
\includegraphics[scale=0.07]{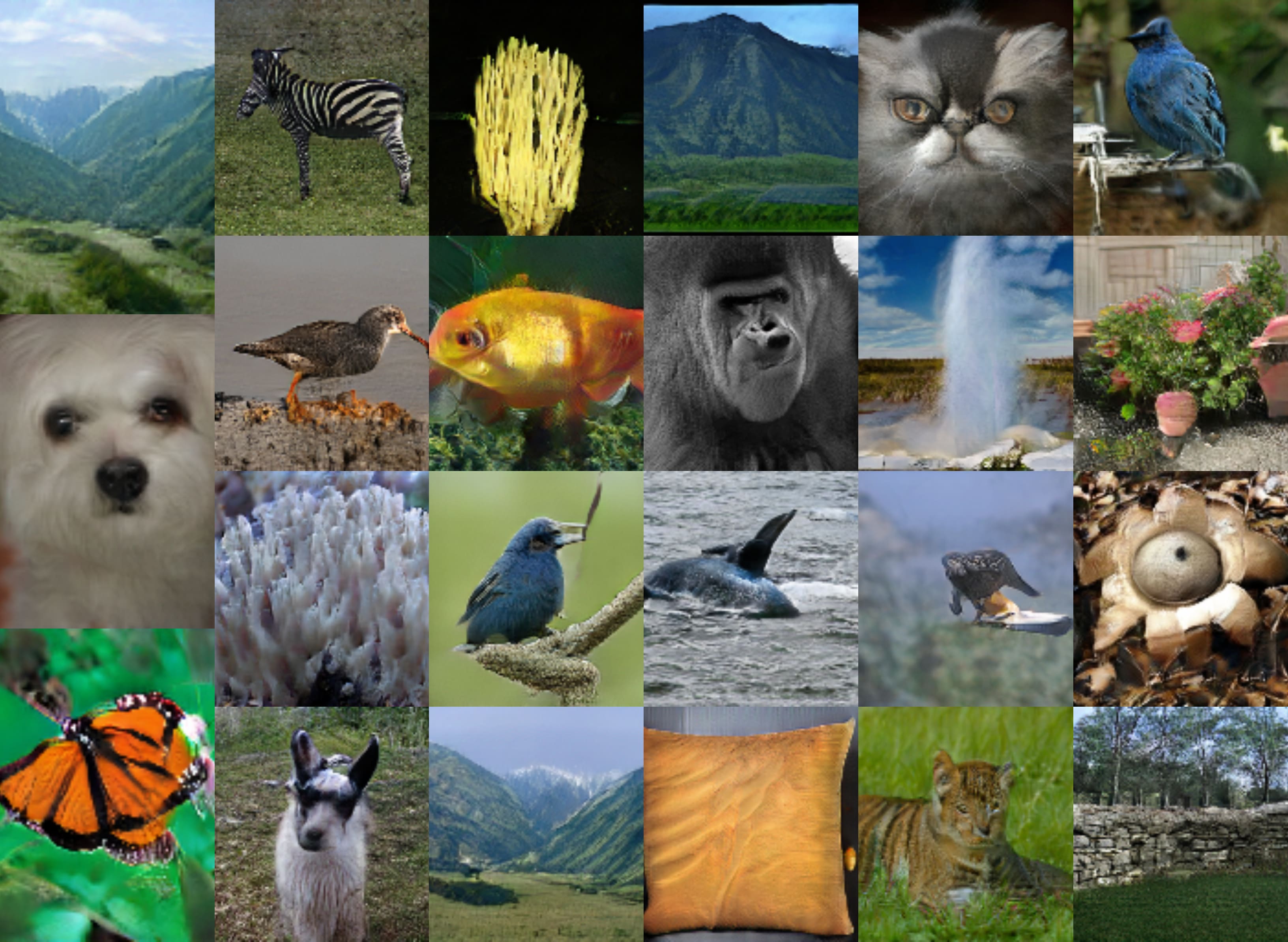}
\end{subfigure}
    \caption{\footnotesize{Upper Panel: YLG conditional image generation on different dog breeds from ImageNet dataset. From up to down: eskimo husky, siberian husky, saint bernard, maltese. Lower Panel: Random generated samples from YLG-SAGAN. Additional generated images are included in the Appendix.}}
\label{samples}
\end{figure}

\section{Acknowledgments}
We would like to wholeheartedly thank the TensorFlow Research Cloud (TFRC) program that gave us access to v3-8 Cloud TPUs and GCP credits to train our models on ImageNet.

{\small
\bibliographystyle{ieee_fullname}
\bibliography{egbib}

\begin{thebibliography}{10}\itemsep=-1pt

\bibitem{ahlswede2000network}
Rudolf Ahlswede, Ning Cai, S-YR Li, and Raymond~W Yeung.
\newblock Network information flow.
\newblock {\em IEEE Transactions on information theory}, 46(4):1204--1216,
  2000.

\bibitem{cannot_generate}
David {Bau}, Jun-Yan {Zhu}, Jonas {Wulff}, William {Peebles}, Hendrik
  {Strobelt}, Bolei {Zhou}, and Antonio {Torralba}.
\newblock {Seeing What a GAN Cannot Generate}.
\newblock {\em arXiv e-prints}, page arXiv:1910.11626, Oct 2019.

\bibitem{bora2017compressed}
Ashish Bora, Ajil Jalal, Eric Price, and Alexandros~G Dimakis.
\newblock Compressed sensing using generative models.
\newblock In {\em Proceedings of the 34th International Conference on Machine
  Learning-Volume 70}, pages 537--546. JMLR. org, 2017.

\bibitem{biggan}
Andrew {Brock}, Jeff {Donahue}, and Karen {Simonyan}.
\newblock {Large Scale GAN Training for High Fidelity Natural Image Synthesis}.
\newblock {\em arXiv e-prints}, page arXiv:1809.11096, Sep 2018.

\bibitem{scram}
Dan~A. {Calian}, Peter {Roelants}, Jacques {Cali}, Ben {Carr}, Krishna {Dubba},
  John~E. {Reid}, and Dell {Zhang}.
\newblock {SCRAM: Spatially Coherent Randomized Attention Maps}.
\newblock {\em arXiv e-prints}, page arXiv:1905.10308, May 2019.

\bibitem{child2019generating}
Rewon Child, Scott Gray, Alec Radford, and Ilya Sutskever.
\newblock Generating long sequences with sparse transformers.
\newblock {\em arXiv preprint arXiv:1904.10509}, 2019.

\bibitem{transformer_xl}
Zihang {Dai}, Zhilin {Yang}, Yiming {Yang}, Jaime {Carbonell}, Quoc~V. {Le},
  and Ruslan {Salakhutdinov}.
\newblock {Transformer-XL: Attentive Language Models Beyond a Fixed-Length
  Context}.
\newblock {\em arXiv e-prints}, page arXiv:1901.02860, Jan 2019.

\bibitem{dimakis2010network}
Alexandros~G Dimakis, P~Brighten Godfrey, Yunnan Wu, Martin~J Wainwright, and
  Kannan Ramchandran.
\newblock Network coding for distributed storage systems.
\newblock {\em IEEE transactions on information theory}, 56(9):4539--4551,
  2010.

\bibitem{donahue2016adversarial}
Jeff Donahue, Philipp Kr{\"a}henb{\"u}hl, and Trevor Darrell.
\newblock Adversarial feature learning.
\newblock {\em arXiv preprint arXiv:1605.09782}, 2016.

\bibitem{gans}
Ian~J. {Goodfellow}, Jean {Pouget-Abadie}, Mehdi {Mirza}, Bing {Xu}, David
  {Warde-Farley}, Sherjil {Ozair}, Aaron {Courville}, and Yoshua {Bengio}.
\newblock {Generative Adversarial Networks}.
\newblock {\em arXiv e-prints}, page arXiv:1406.2661, Jun 2014.

\bibitem{BLOCKSPARSE}
Scott Gray, Alec Radford, and Diederik~P Kingma.
\newblock Gpu kernels for block-sparse weights.
\newblock {\em arXiv preprint arXiv:1711.09224}, 2017.

\bibitem{star}
Qipeng {Guo}, Xipeng {Qiu}, Pengfei {Liu}, Yunfan {Shao}, Xiangyang {Xue}, and
  Zheng {Zhang}.
\newblock {Star-Transformer}.
\newblock {\em arXiv e-prints}, page arXiv:1902.09113, Feb 2019.

\bibitem{hand2019global}
Paul Hand and Vladislav Voroninski.
\newblock Global guarantees for enforcing deep generative priors by empirical
  risk.
\newblock {\em IEEE Transactions on Information Theory}, 2019.

\bibitem{kabkab2018task}
Maya Kabkab, Pouya Samangouei, and Rama Chellappa.
\newblock Task-aware compressed sensing with generative adversarial networks.
\newblock In {\em Thirty-Second AAAI Conference on Artificial Intelligence},
  2018.

\bibitem{stylegan}
Tero {Karras}, Samuli {Laine}, and Timo {Aila}.
\newblock {A Style-Based Generator Architecture for Generative Adversarial
  Networks}.
\newblock {\em arXiv e-prints}, page arXiv:1812.04948, Dec 2018.

\bibitem{lipton2017precise}
Zachary~C Lipton and Subarna Tripathi.
\newblock Precise recovery of latent vectors from generative adversarial
  networks.
\newblock {\em arXiv preprint arXiv:1702.04782}, 2017.

\bibitem{imagetransformer}
Niki Parmar, Ashish Vaswani, Jakob Uszkoreit, {\L}ukasz Kaiser, Noam Shazeer,
  Alexander Ku, and Dustin Tran.
\newblock Image transformer.
\newblock {\em arXiv preprint arXiv:1802.05751}, 2018.

\bibitem{dcgan}
Alec {Radford}, Luke {Metz}, and Soumith {Chintala}.
\newblock {Unsupervised Representation Learning with Deep Convolutional
  Generative Adversarial Networks}.
\newblock {\em arXiv e-prints}, page arXiv:1511.06434, Nov 2015.

\bibitem{raj2019gan}
Ankit Raj, Yuqi Li, and Yoram Bresler.
\newblock Gan-based projector for faster recovery with convergence guarantees
  in linear inverse problems.
\newblock In {\em Proceedings of the IEEE International Conference on Computer
  Vision}, pages 5602--5611, 2019.

\bibitem{rick2017one}
JH Rick~Chang, Chun-Liang Li, Barnabas Poczos, BVK Vijaya~Kumar, and Aswin~C
  Sankaranarayanan.
\newblock One network to solve them all--solving linear inverse problems using
  deep projection models.
\newblock In {\em Proceedings of the IEEE International Conference on Computer
  Vision}, pages 5888--5897, 2017.

\bibitem{ILSVRC15}
Olga Russakovsky, Jia Deng, Hao Su, Jonathan Krause, Sanjeev Satheesh, Sean Ma,
  Zhiheng Huang, Andrej Karpathy, Aditya Khosla, Michael Bernstein,
  Alexander~C. Berg, and Li Fei-Fei.
\newblock {ImageNet Large Scale Visual Recognition Challenge}.
\newblock {\em International Journal of Computer Vision (IJCV)},
  115(3):211--252, 2015.

\bibitem{vgg}
Karen {Simonyan} and Andrew {Zisserman}.
\newblock {Very Deep Convolutional Networks for Large-Scale Image Recognition}.
\newblock {\em arXiv e-prints}, page arXiv:1409.1556, Sep 2014.

\bibitem{song2019surfing}
Ganlin Song, Zhou Fan, and John Lafferty.
\newblock Surfing: Iterative optimization over incrementally trained deep
  networks.
\newblock {\em arXiv preprint arXiv:1907.08653}, 2019.

\bibitem{adaptive_span}
Sainbayar {Sukhbaatar}, Edouard {Grave}, Piotr {Bojanowski}, and Armand
  {Joulin}.
\newblock {Adaptive Attention Span in Transformers}.
\newblock {\em arXiv e-prints}, page arXiv:1905.07799, May 2019.

\bibitem{attention_is_all_you_need}
Ashish {Vaswani}, Noam {Shazeer}, Niki {Parmar}, Jakob {Uszkoreit}, Llion
  {Jones}, Aidan~N. {Gomez}, Lukasz {Kaiser}, and Illia {Polosukhin}.
\newblock {Attention Is All You Need}.
\newblock {\em arXiv e-prints}, page arXiv:1706.03762, Jun 2017.

\bibitem{sagan}
Han {Zhang}, Ian {Goodfellow}, Dimitris {Metaxas}, and Augustus {Odena}.
\newblock {Self-Attention Generative Adversarial Networks}.
\newblock {\em arXiv e-prints}, page arXiv:1805.08318, May 2018.

\bibitem{stackgan}
Han {Zhang}, Tao {Xu}, Hongsheng {Li}, Shaoting {Zhang}, Xiaogang {Wang},
  Xiaolei {Huang}, and Dimitris {Metaxas}.
\newblock {StackGAN: Text to Photo-realistic Image Synthesis with Stacked
  Generative Adversarial Networks}.
\newblock {\em arXiv e-prints}, page arXiv:1612.03242, Dec 2016.

\bibitem{lookahead}
Michael~R. {Zhang}, James {Lucas}, Geoffrey {Hinton}, and Jimmy {Ba}.
\newblock {Lookahead Optimizer: k steps forward, 1 step back}.
\newblock {\em arXiv e-prints}, page arXiv:1907.08610, Jul 2019.

\end{thebibliography}
}

\section{Appendix}
\subsection{A closer look to our inversion method}
\label{closer_look}
This subsection aims to explain technical details of our inversion technique and clarify the details of our approach. 

We begin with a recap of our method.
Given a real image we pass it to the discriminator and we extract the attention map from the attention layer. This attention map contains for every point of the query image, a probability distribution over the pixels of the key image. We can then convert this attention map to a saliency map: by averaging the attention each key point gets from all the query points, we can get a probability distribution over the "importance" of the pixels of the key image. We denote this saliency map with $S$.
Our proposed inversion algorithm is to perform gradient descent to minimize the discriminator embedding distance, weighted by this salience map: 
$$
            \lVert \left( D^0(G(z)) - D^0(x) \right) \cdot S' \rVert ^2,
$$
where $S'$ is a projected version of saliency map $S$, $x$ is the image, and $D^0$ is the Discriminator network up to, but not including, the attention layer.

\subsubsection{Multiple heads and saliency map}
There are some practical considerations that we need to address before illustrating that our inversion method indeed works: the most important of which is how the saliency map $S$ looks like.

In our analysis of the YLG attention layers, we explain that because of the Full Information property, our patterns are able, potentially, to discover a dependency between any two pixels of an image. If that is true, we should expect that in the general case our saliency map, generated by the average of all heads, allocates non-zero weights to all image pixels. The important question becomes whether this joint saliency map weights more the pixels that are important for a visually convincing inversion. For example, in case of a bird flying with a blue-sky in the background, we should be ready to accept a small error in some point in the clouds of the sky but not a bird deformation that will make the inverted image look unrealistic. Therefore, our saliency map should allocate more weight in the bird than in it allocates in the background sky.

\begin{figure*}
    \begin{subfigure}{0.3\textwidth}
    \centering
        \includegraphics[scale=0.82]{figures/redshank/redshank.png}
        \caption{}
    \end{subfigure}
    \begin{subfigure}{0.3\textwidth}
    \centering
        \includegraphics[scale=0.4]{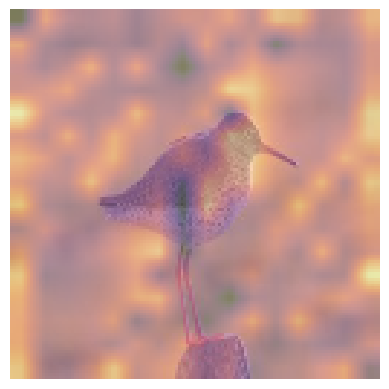}
        \subcaption{}
        \label{all_saliency}
    \end{subfigure}
    \begin{subfigure}{0.3\textwidth}
    \centering
        \includegraphics[scale=0.4]{figures/redshank/saliency/saliency_7.png}
        \subcaption{}
        \label{single_saliency}
    \end{subfigure}
    \caption{\footnotesize{(a) Real image of a redshank. (b) Saliency map extracted from \textbf{all} heads of the Discriminator. (c) Saliency map extracted from a \textbf{single} head of the Discriminator. Weighting our loss function with (b) does not have a huge impact, as the attention weights are almost uniform. Saliency map from (c) is more likely to help correct inversion of the bird. We can use saliency maps from other heads to invert the background as well.}}
\end{figure*}

We already showed in Section \ref{inversion_lens} that different heads specialize in discovering important image parts (for example,
some heads learn to focus on local neighbhoords, important shape edges, background, etc.) so extracting a saliency map $S$ by averaging all heads usually leads in a uniform distribution over pixels, which is not helping inversion. Figure \ref{all_saliency} shows the saliency map jointly all heads of the attention layer of the discriminator produce. Although the bird receives a bit more attention than the background, it is not clear how this map would help weight our loss for inversion. However, as illustrated in \ref{single_saliency}, there are heads that produce far more meaningful saliency maps for a good-looking inversion. There is a drawback here as well though; if we use that head only, we completely miss the background.

To address this problem, we find two solutions that work quite well.
\begin{itemize}
    \item Solution 1: calculate Equation \ref{disc_loss} separately for each head and then add the losses. In that case, the new loss function is given by the following equation: 
    \begin{equation}
            \sum_{i} \lVert \left( D^0(G(z)) - D^0(x) \right) \cdot S_i' \rVert ^2,
            \label{disc_head_loss}
\end{equation}
where $S_i'$ is the saliency map extracted from head $i$.
    \item Solution 2: Examine manually the saliency maps for each head and remove the heads that are attending mainly to non-crucial for the inversion areas, such as homogeneous backgrounds.
\end{itemize}

\subsubsection{More inversion visualizations}
\begin{figure*}
    \centering
    \begin{subfigure}{0.45\textwidth}
        \includegraphics[scale=0.07]{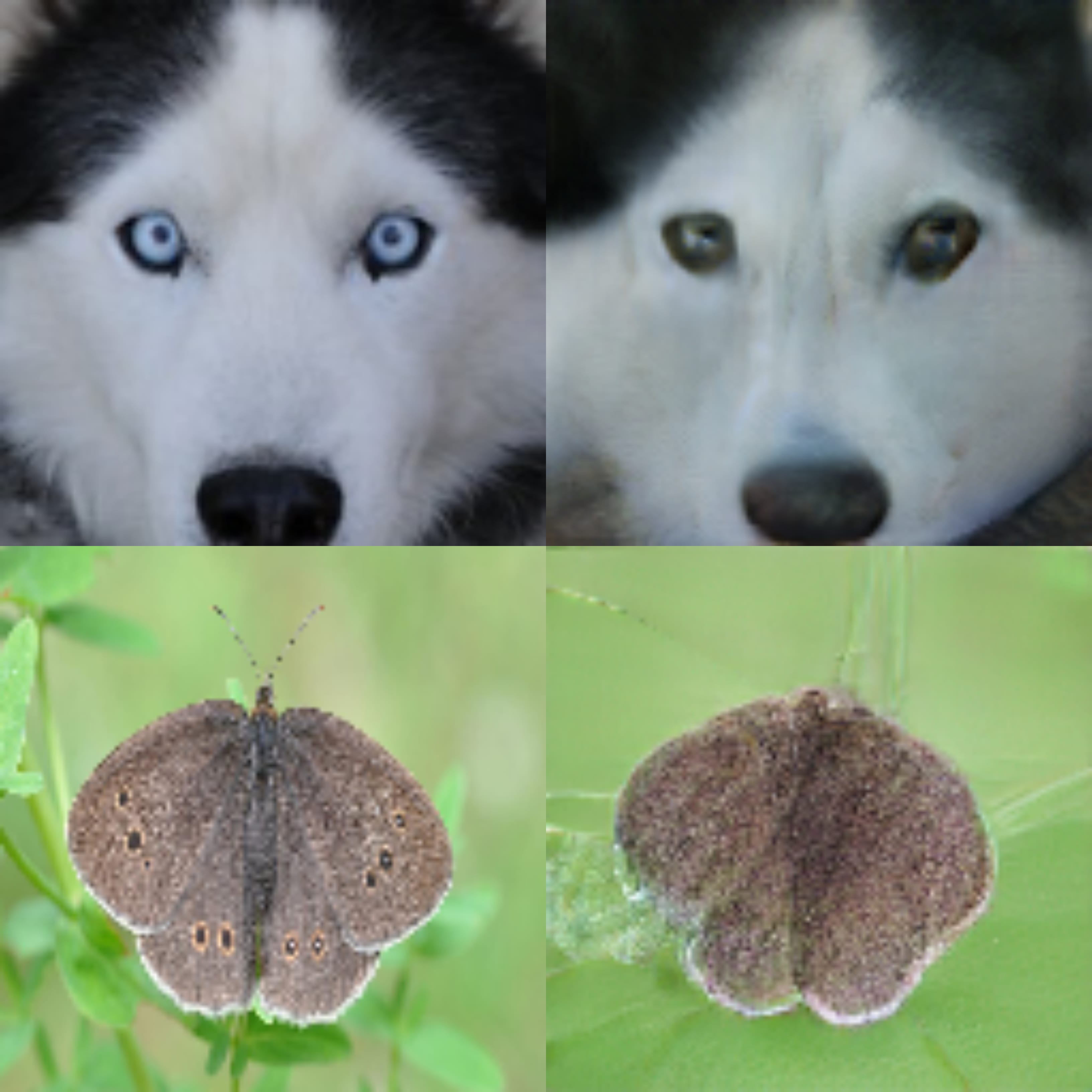}
    \end{subfigure}
    \begin{subfigure}{0.45\textwidth}
        \includegraphics[scale=0.07]{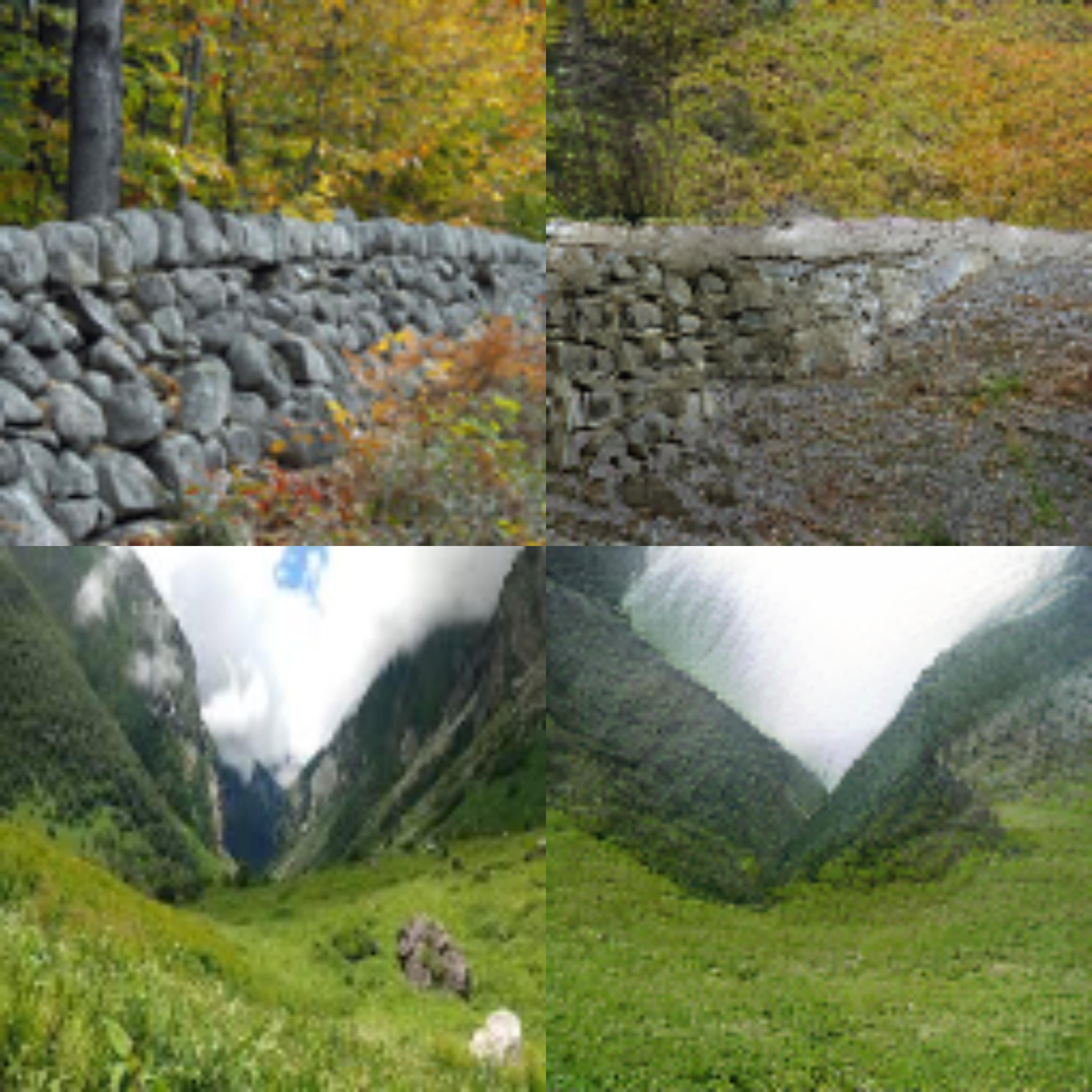}
    \end{subfigure}
    
    \begin{subfigure}{0.45\textwidth}
        \includegraphics[scale=0.07]{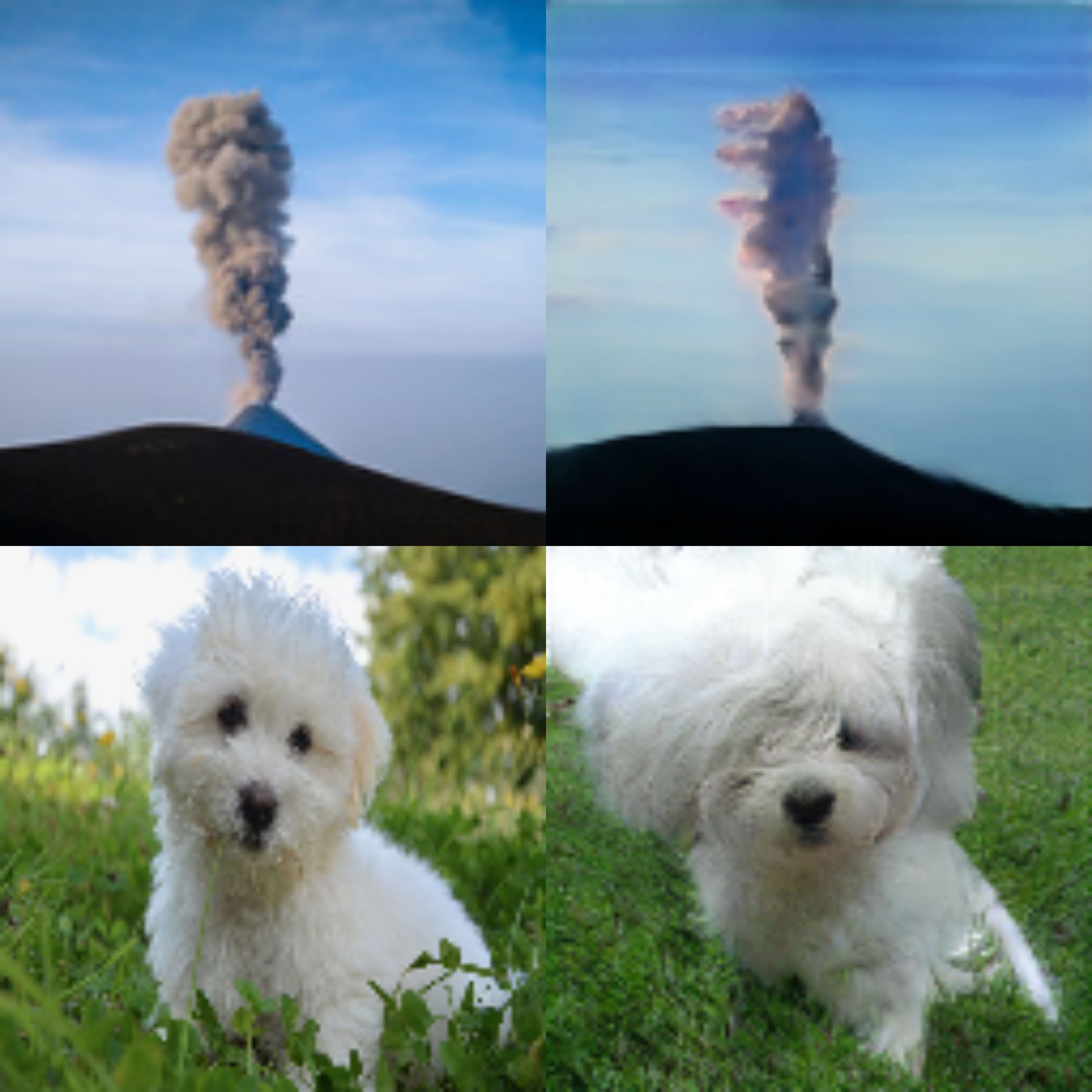}
    \end{subfigure}
    \begin{subfigure}{0.45\textwidth}
        \includegraphics[scale=0.07]{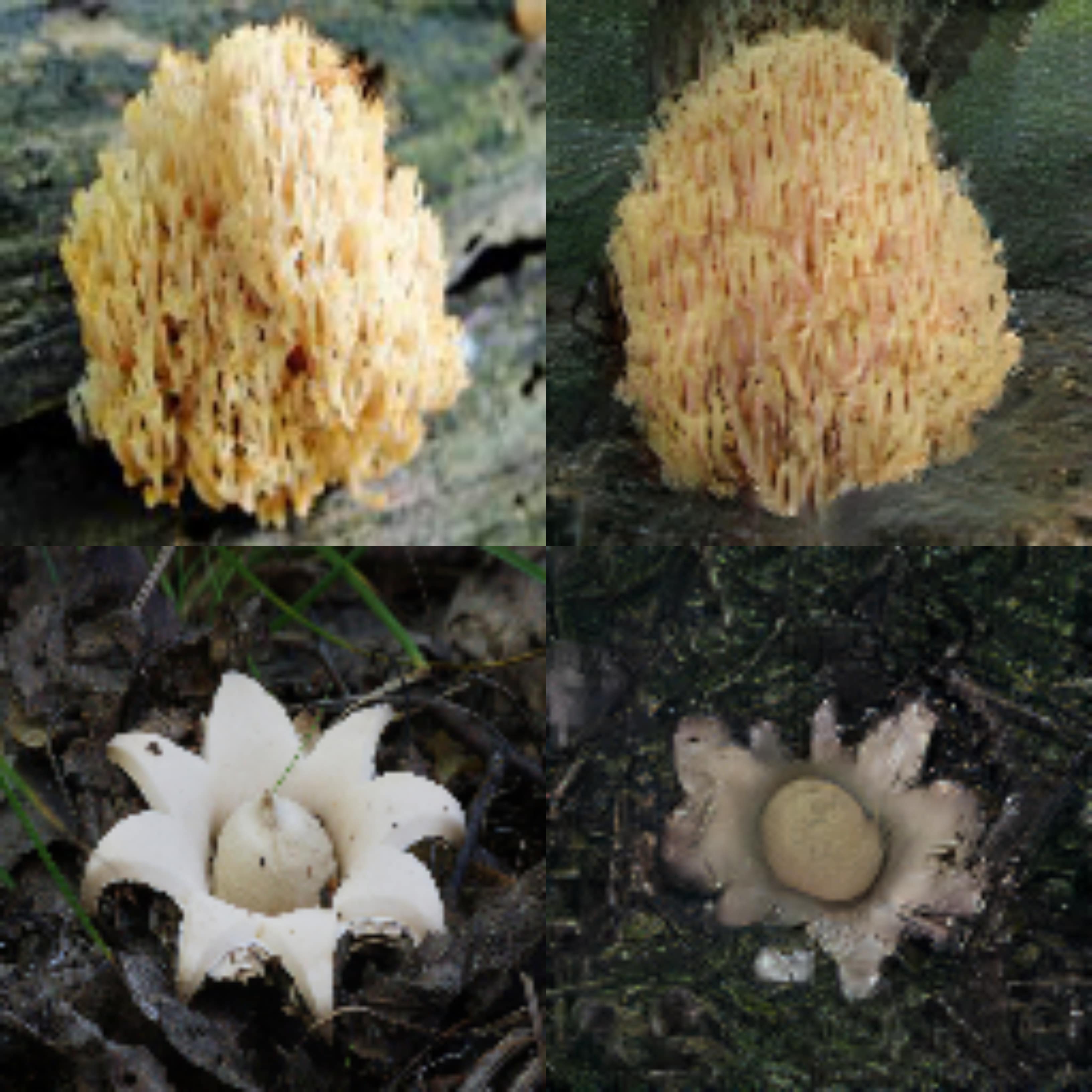}
    \end{subfigure}
\caption{More inversions using our technique. To the left we present real images and to the right our inversions using YLG SAGAN.}
\label{inversions}
\end{figure*}

We present several inversions for different categories of real images at Figure \ref{inversions}. In all our Figures, we use Solution 1 as it has the advantage that it does not require human supervision.

With our method, we can effectively invert real world scenes. We tested the standard inversion method~\cite{bora2017compressed} for these images as well and the results were far less impressive for all images. Especially for the dogs, we noted complete failure of the previous approach, similar to what we illustrate in Figure \ref{fig:husky_inv}.

\subsubsection{Experiments setup}
In this subsection, we will briefly describe the experimental setup for our inversion technique. We choose to use the recently introduced Lookahead~\cite{lookahead} optimizer as we find that it reduces the number of different seeds we have to try for a successful inversion. For the vast majority of the examined real images, we are able to get a satisfying inversion by trying at most 4 different seeds. We set the learning rate to $0.05$ and we update for maximum $1500$ steps. On a single V100 GPU, a single image inversion takes less than a minute to complete. We choose to invert real-world images that were not present in the training set. We initialize our latent variables from a truncated normal distribution, as explained in \ref{truncation}.

\subsection{Truncation and how it helps inversion}
\label{truncation}
In the BigGAN~\cite{biggan} paper, the authors observed that latent variables sampled from a truncated normal distribution generated generally more photo-realistic images compared to ones generated from the normal distribution which was used during the training.  This so-called truncation trick (resampling the values with magnitude above a chosen threshold) leads to improvement in sample quality at the cost of reduction in sample variety. For the generated images of YLG presented in this paper, we also utilized this trick.

Interestingly, the truncation trick can help inversion as well under some conditions. If the original image has good quality, then according to the truncation trick, it is more probable to be generated by a latent variable sampled from a truncated normal (where values which fall outside a range are resampled to fall inside that range) than the standard normal distribution $N(0, I)$. For that reason, in our inversions we start our trainable latent variable from a sample of the truncated normal distribution. We found experimentally that setting the truncation threshold to two standard deviations from the median (in our case 0), is a good trade-off between producing photo-realistic images and having enough diversity to invert an arbitrary real world image.

\subsection{Strided Pattern}
In the ablation studies of our paper, we train a model we name YLG - Strided. For this model, we report better results than the baseline SAGAN~\cite{sagan} model and slightly worse results than the proposed YLG model.
The purpose of this section is to give more information on how YLG and YLG - Strided differ.

First of all, the only difference between YLG and YLG Strided is the choosing of attention masks for the attention heads: both models implement 2-step attention patterns with Full Information and two-dimensional locality using the ESA framework.

YLG model uses the RTL and LTR patterns introduced in the paper (see Figures \ref{RTL_mask}, \ref{LTR_mask}). Each pattern corresponds to a two-step attention: in our implementation of multi-step attention we compute steps in parallel using multiple heads, so in total we need 8 attention heads for YLG. In YLG - Strided instead of using different patterns (RTL and LTR), we stick with using a single attention pattern.
Our motivation is to: (i) investigate whether using multiple attention patterns simultaneously affects performance, (ii) discover whether the performance differences between one-dimensional sparse patterns reported in the literature remain when the patterns are rendered to be aware of two-dimensional geometry. To explore (i), (ii) a natural choice was to work with the Strided pattern proposed in Sparse Transformers~\cite{child2019generating} as it was found to be (i) effective for modeling images and (ii) more suitable than the Fixed pattern (see Figure 2a), on which we built to invent LTR, RTL. 

\begin{figure*}[!htb]
    \begin{center}
            \begin{subfigure}{0.45\textwidth}
            \centering
            \includegraphics[scale=0.75]{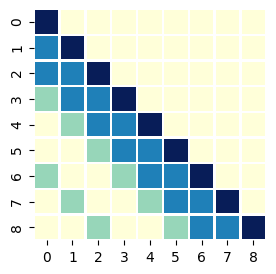}
            \subcaption{\footnotesize{Attention masks for Strided Pattern ~\cite{child2019generating}.} }
            \label{NSTRIDED_mask}
        \end{subfigure}
        \quad
        \begin{subfigure}{0.45\textwidth}
            \centering
            \includegraphics[scale=0.75]{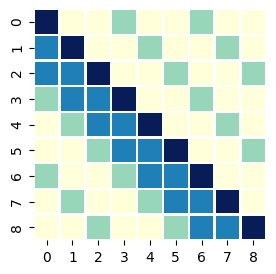}
            \subcaption{\footnotesize{Attention masks for YLG - Strided (Extended Strided with Full Information property)}}
            \label{STRIDED_mask}
        \end{subfigure}
        
        \begin{subfigure}{0.45\textwidth}
            \centering
            \includegraphics[scale=0.75]{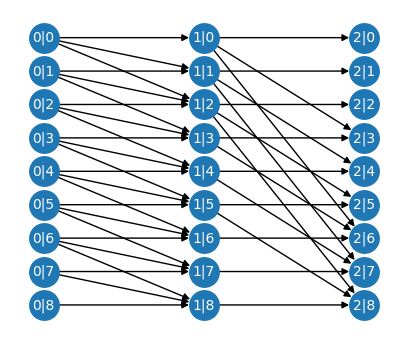}
            \subcaption{\footnotesize{Information Flow Graph associated with Strided Pattern. This pattern \textit{does not have Full Information}, i.e. there are dependencies between nodes that the attention layer cannot model. For example, there is no path from node $2$ of $V^0$ to node $1$ of $V^2$.}}
            \label{NSTRIDED_IFG}
        \end{subfigure}
        \quad
        \begin{subfigure}{0.45\textwidth}
            \centering
            \includegraphics[scale=0.75]{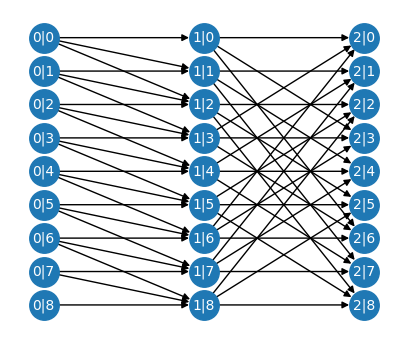}
            \subcaption{
            \footnotesize{
            Information Flow Graph associated with YLG - Strided pattern. This pattern has \textbf{Full Information}, i.e. there is a path between any node of $V^0$ and any node of $V^2$. Note that the number of edges is only increased by a constant compared to the Strided Attention Pattern~\cite{child2019generating}, illustrated in \ref{NSTRIDED_mask}.}}
            \label{STRIDED_IFG}
        \end{subfigure}
    \end{center}
    \caption{\small{This Figure illustrates the original Strided Pattern~\cite{child2019generating} and the YLG - Strided pattern which has Full Information. First row demonstrates the different boolean masks that we apply to each of the two steps. Color of cell [i. j] indicates whether node i can attend to node j. With dark blue we indicate the attended positions in both steps. With light blue the positions of the first mask and with green the positions of the second mask. The yellow cells correspond to positions that we do not attend to any step (sparsity). The second row illustrates Information Flow Graph associated with the aforementioned attention masks. An Information Flow Graph visualizes how information "flows" in the attention layer. Intuitively, it visualizes how our model can use the 2-step factorization to find dependencies between image pixels. At each multipartite graph, the nodes of the first vertex set correspond to the image pixels, just before the attention. An edge from a node of the first vertex set, $V^0$, to a node of the second vertex set, $V^1$, means that the node of $V^0$ can attend to node of $V^1$ at the first attention step. Edges between $V^1, V^2$ illustrate the second attention step.}}
    \label{strided_pattn}
\end{figure*}

We illustrate the Strided pattern, as proposed in Sparse Transformers~\cite{child2019generating}, in Figures \ref{NSTRIDED_mask}, \ref{NSTRIDED_IFG}. For a fair comparison with LTR, RTL we need to expand Strided pattern in order for it to have Full Information. Figures \ref{STRIDED_mask}, \ref{STRIDED_IFG} illustrate this expansion. The pattern illustrated in this Figure is exactly the pattern that YLG - Strided uses. Note that this pattern attends to the same order of positions, $O(n\sqrt n)$, as LTR and RTL. For one to one comparison with YLG, YLG - Strided has also 8 heads: the Full Information pattern is implemented 4 times, as we need 2 heads for a 2-step pattern. As already mentioned, we also use ESA framework for YLG - Strided.

\subsection{Things that did not work}
In this subsection, we present several ideas, relevant to the paper, that we experimented on and found that their results were not satisfying. Our motivation is to inform the research community about the observed shortcomings of these approaches so that other researchers can re-formulate them, reject them or compare their findings with ours.

\subsubsection{Weighted inversion at the generator space}

\begin{figure*}
    \begin{subfigure}{0.24\textwidth}
    \centering
    \includegraphics[scale=0.6]{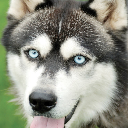}
    \subcaption{Real image.}
    \label{real_husky}
    \end{subfigure}
    \begin{subfigure}{0.24\textwidth}
    \centering
    \includegraphics[scale=0.6]{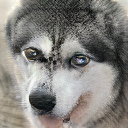}
    \subcaption{Inversion with our method.}
    \label{husky_ours}
    \end{subfigure}
    \begin{subfigure}{0.24\textwidth}
    \centering
    \includegraphics[scale=0.6]{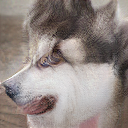}
    \subcaption{Weighted inversion at Generator.}
    \label{gen_husky}
    \end{subfigure}
    \begin{subfigure}{0.24\textwidth}
    \centering
    \includegraphics[scale=0.6]{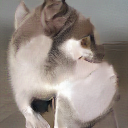}
    \subcaption{Inversion using the standard method~\cite{bora2017compressed}.}
    \label{bora_husky}
    \end{subfigure}
    \caption{Inversion with different methods of the real image of \ref{real_husky}. Our method, \ref{husky_ours}, is the only successful inversion. The inversion using the weights from the saliency map to the output of the Generator, \ref{gen_husky}, fails badly. The same holds for inversion using the standard method in the literature~\cite{bora2017compressed}, as shown in \ref{bora_husky}. }
    \label{fig:husky_inv}
\end{figure*}
We already discussed that our key idea for the inversion:
we pass a real image to the discriminator, extract the attention map, convert the attention map to a saliency distribution $S$ and we perform gradient descent to minimize the discriminator embedding distance, weighted by this saliency map:
$$
\lVert \left( D^0(G(z)) - D^0(x) \right) \cdot S' \rVert ^2,
$$
where $S'$ is a projected version of saliency map $S$, $x$ is the image, and $D^0$ is the Discriminator network up to, but not including, the attention layer. In practise, we use Equation \ref{disc_head_loss} for the reasons we explained in Section \ref{closer_look} but for the rest of this Section we will overlook this detail as it is not important for our point.

Equation \ref{disc_loss} implies that the inversion takes place in the embedding space of the Discriminator. However, naturally one might wonder if we could use the saliency map $S$ to weight the inversion of the Generator, in other words, if we could perform gradient descent on:
\begin{equation}
            \lVert \left( G(z) - x \right) \cdot S'' \rVert ^2,
            \label{gen_loss}
\end{equation}
where $S''$ is a projected version of $S$ to match the dimensions of the Generator network.

In our experiments, we find that this approach generally leads to inversions of poor quality. To illustrate this, we present inversions of an image of a real husky from the the weighted generator inversion, the weighted discriminator inversion and standard inversion method~\cite{bora2017compressed} at Figure \ref{fig:husky_inv}.

There are several reasons that could explain the quality gap when we change from inversion to the space of the Discriminator to that of the Generator. First of all, the saliency map we use to weight our loss is extracted from the Discriminator, which means that the weights reflect what the Discriminator network considers important at that stage. Therefore, it is reasonable to expect that this saliency map would be more accurate to describe what is important for the input of the attention of the discriminator than to the output of the Generator. Also note that due to the layers of the Discriminator before the attention, the images of the output of the generator and the input of the attention of the Discriminator can be quite different. Finally, the Discriminator may provide an "easier" embedding space for inversion. The idea of using a different embedding space than the output of the Generator it is not new; activations from VGG16~\cite{vgg} have also been used for inversion~\cite{cannot_generate}. Our novelty is that we use the Discriminator instead of another pre-trained model to work on a new embedding space.

\subsubsection{Combination of dense and sparse heads}
In our paper, we provide strong experimental evidence that multi-step two-dimensional sparse local heads can be more efficient than the conventional dense attention layer. We justify this evidence theoretically by modelling the multi-step attention with Information Flow Graphs and indicating the implications of Full Information. Naturally, one might wonder what would happen if we combine YLG attention with dense attention. To answer this question, we split heads into two groups, the local - sparse heads and the dense ones. Specifically, we use 4 heads that implement the RTL, LTR patterns (see paper for more details) and 4 dense heads and we train this variation of SAGAN. We use the same setup as with our other experiments. We report FID 19.21 and Inception: 51.23. These scores are far behind than the scores of YLG and thus we did not see any benefit continuing the research in this direction.

\subsubsection{Different resolution heads}
One idea we believed it would be interesting was to train SAGAN with a multi-headed dense attention layer of different resolution heads. In simple words, that means that in this attention layer some heads have a wider vector representation than others. Our motivation was that the different resolutions could have helped enforcing locality in a different way; we expected the heads with the narrow hidden representations to learn to attend only locally and the wider heads to be able to recover long-range dependencies. 

In SAGAN, the number of channels in the query vector is 32, so for an 8-head attention layer normally each head would get 4 positions. We split the 8 heads into two equal groups: the narrow and the wide heads. In our experiment, narrow heads get only 2 positions for their vector representation while wide heads get 6. After training on the same setup with our other experiments, we obtain FID 19.57 and Inception score: 50.93. These scores are slightly worse than the original SAGAN, but are far better than SAGAN with dense 8-head attention which achieved FID 20.09 and Inception 46.01, as mentioned in the ablation study. 

At least in our preliminary experiments, different resolution heads were not found to help very much. Perhaps they can be combined with YLG attention but we more research would be needed in this direction.

\subsection{Information Flow Graphs} 
We found that thinking about sparse attention as a network with multiple stages is helpful in visualizing how information of different tokens is attended and combined. 
We use Information Flow Graphs (IFGs) that were introduced in~\cite{dimakis2010network} for modeling how distributed storage codes preserve data. In full generality, IFGs are directed acyclic graphs with capacitated directed edges. Each storage node is represented with two copies of a vertex ($x_\text{in}$ and $x_\text{out}$) connected by a directed edge with capacity equal to the amount of information that can be stored into that node. The key insight is that a multi-stage attention network can be considered a storage network since intermediate tokens are representing combinations of tokens at the previous stage. The IFGs we use in this paper are a special case: every token of every stage of an attention layer is represented by a storage node. Since all the tokens have the same size, we can eliminate vertex splitting and compactly represent each storage node by a single vertex, as shown in Figure \ref{STRIDED_IFG}. 

Full information is a design requirement that we found to be helpful in designing attention networks. It simply means that any single input token is connected with a directed path to any output token and hence information (of entropy equal to one token representation) can flow from any one input into any one output. As we discussed in the paper, we found that previously used sparse attention patterns did not have this property and we augmented them to obtain the patterns we use. 
A stronger requirement would be that any \textit{pair} of input nodes is connected to any pair of output nodes with two edge-disjoint paths. This would mean that flow of two tokens can be supported from any input to any output. Note that a fully connected network can support this for any pair or even for any set of $k$ input-output pairs for $\forall k \leq n$. 

An interesting example is the star transformer~\cite{star} where all $n$ input tokens are connected to a single intermediate node which is then connected to all output tokens. This information flow graph has $2n$ directed edges and can indeed support full information. However, it cannot support a flow of $2$ tokens for any pair, since there is a bottleneck at the intermediate node. We believe that enforcing good information flow for pairs or higher size sets improves the design of attention networks and we plan to investigate this further in the future. 

\subsection{Generated images}
\begin{figure*}
\centering
\includegraphics[scale=0.15]{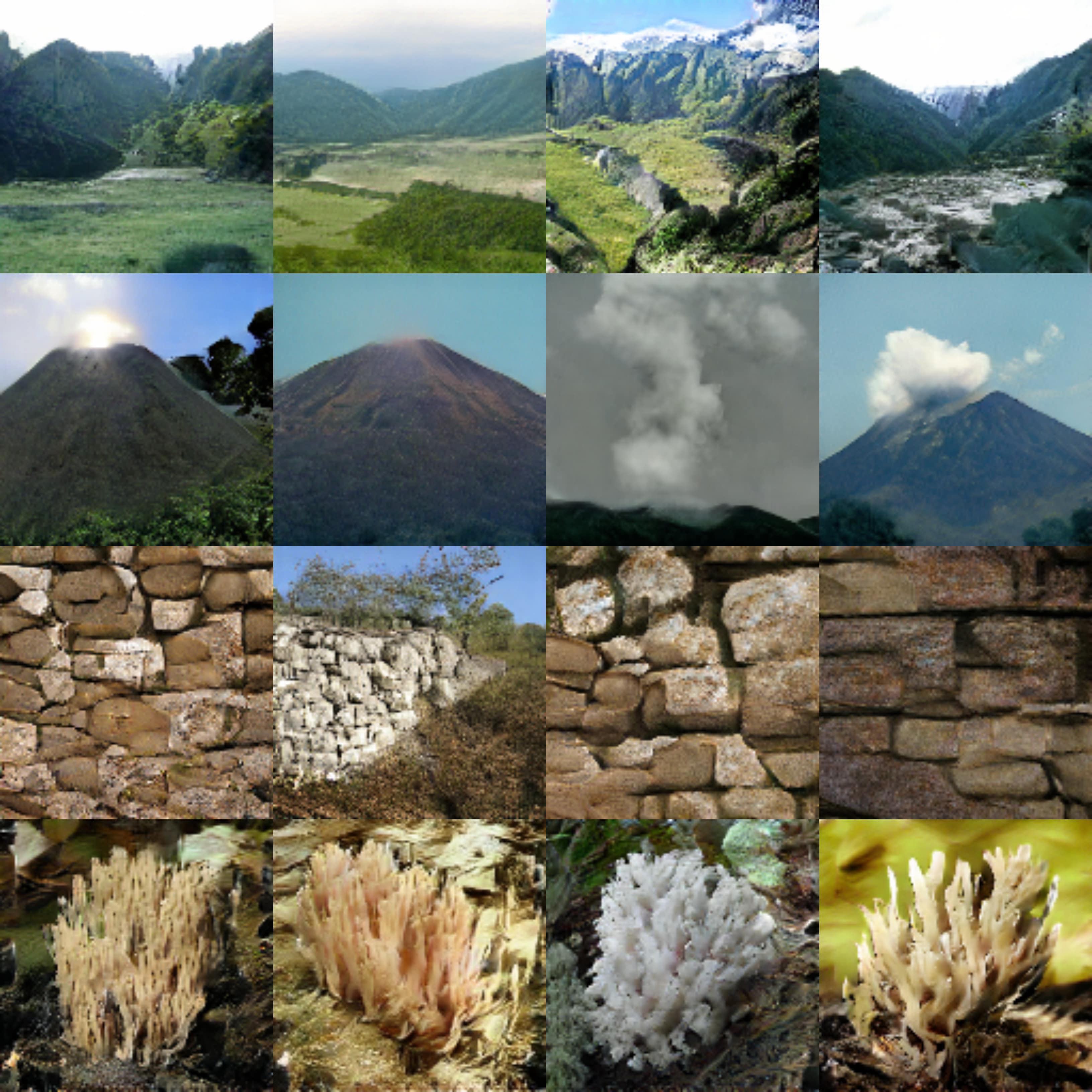}
\caption{Generated images from YLG SAGAN divided by ImageNet category.}
\label{collage1}
\end{figure*}

\begin{figure*}
\centering
\includegraphics[scale=0.15]{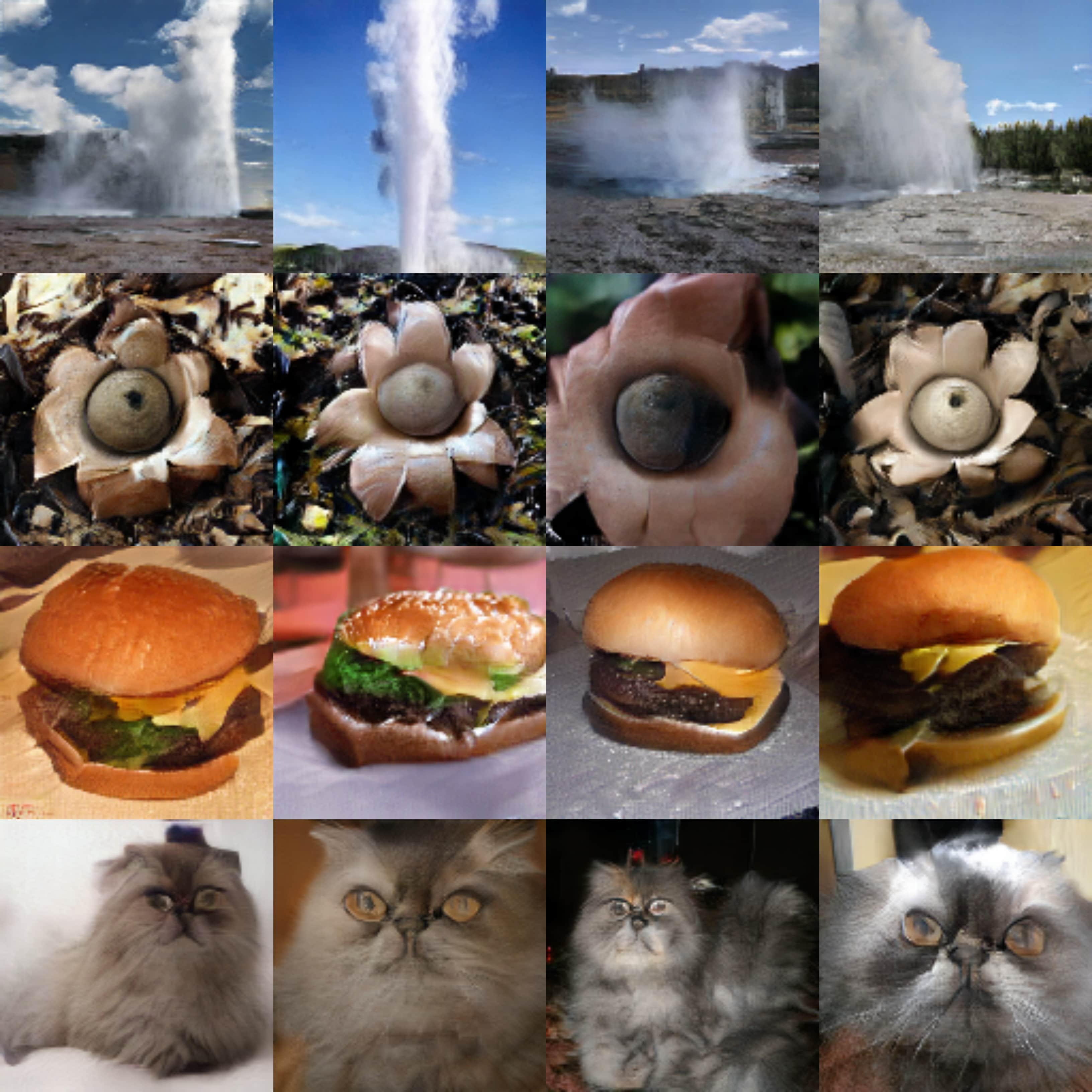}
\caption{Generated images from YLG SAGAN divided by ImageNet category.}
\label{collage2}
\end{figure*}

\begin{figure*}
\centering
\includegraphics[scale=0.15]{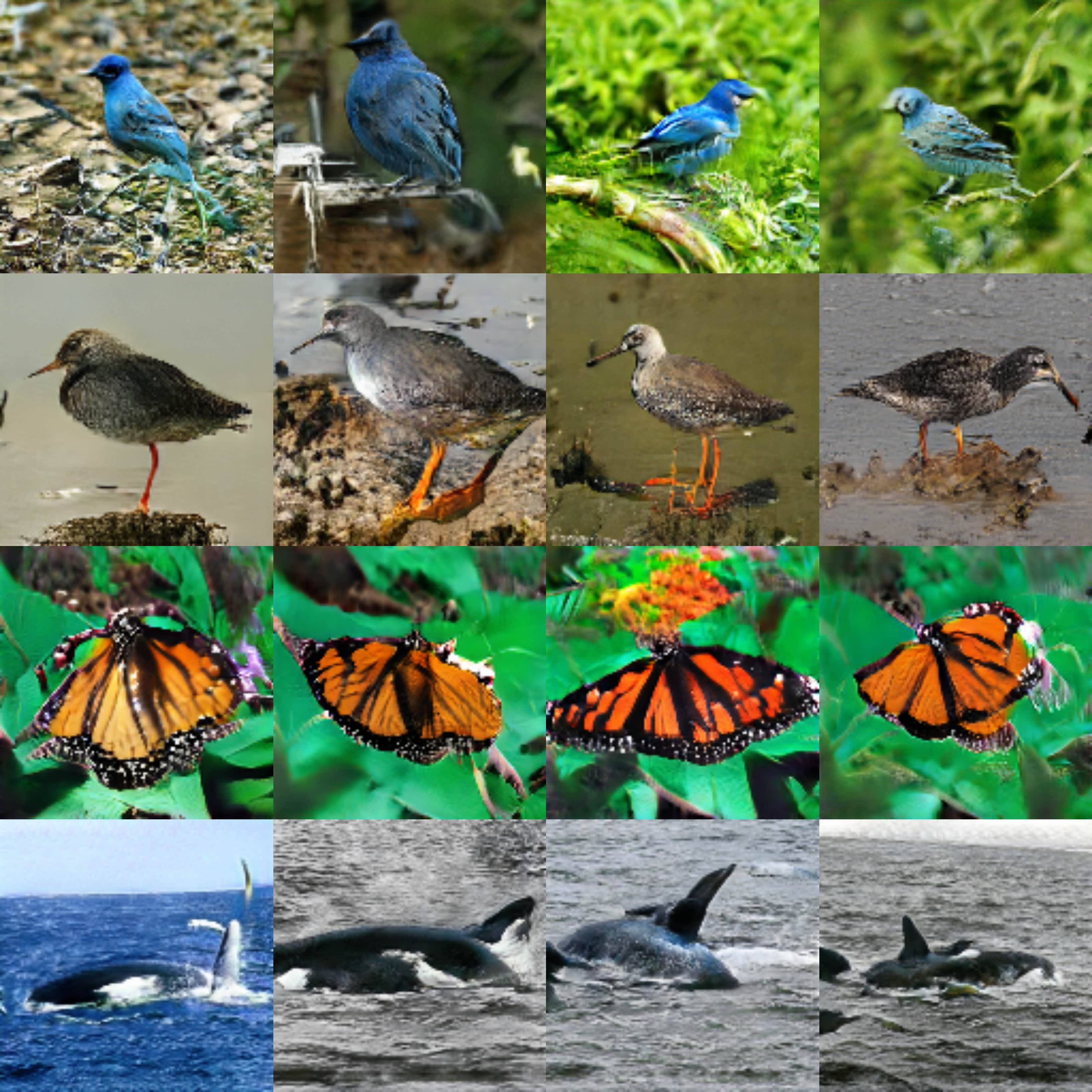}
\caption{Generated images from YLG SAGAN divided by ImageNet category.}
\label{collage3}
\end{figure*}

We present some more generated images from our YLG SAGAN. The images are divided per category and are presented in Figures \ref{collage1}, \ref{collage2}, \ref{collage3}.

\end{document}